\documentclass[10pt,journal,compsoc]{IEEEtran}

\ifCLASSOPTIONcompsoc
  \usepackage[nocompress]{cite}
\else
  \usepackage{cite}
\fi

\ifCLASSINFOpdf
\else
\fi

\usepackage[caption=false,font=normalsize,labelfont=sf,textfont=sf]{subfig}
\usepackage{textcomp}
\usepackage{stfloats}
\usepackage{url}
\usepackage{verbatim}
\usepackage{graphicx}
\usepackage{cite}
\usepackage{amsfonts}
\usepackage{amssymb}
\usepackage{amsthm,amsmath}
\usepackage{mathrsfs}
\usepackage{indentfirst}
\usepackage{multirow}
\usepackage{inconsolata}
\usepackage{microtype}
\usepackage{subcaption}
\usepackage{bbold}
\usepackage{longtable}
\usepackage{arydshln}
\usepackage{array}
\usepackage{booktabs}
\usepackage{wrapfig}
\newcommand\ours{\textsc{PerMU}}
\newcommand\oursfast{$\ours^\dag$}
\newcommand\oursfastdis{$\ours^\dag_{dis}$}

\newcommand\metric{\textsc{MSM}}
\newcommand\ttsmall[1]{\texttt{\textmd {#1}}}
\DeclareOldFontCommand{\rm}
\usepackage{makecell}   
\usepackage{mdframed}
\usepackage{enumitem}
\usepackage[super]{nth}
\usepackage[flushleft]{threeparttable}
\usepackage{xcolor}
\usepackage[numbers]{natbib}
\definecolor{lightpurple}{RGB}{221, 160, 221}
\definecolor{lightblue}{RGB}{173, 216, 230}
\theoremstyle{plain}
\newtheorem{theorem}{Theorem}[section]
\newtheorem{proposition}[theorem]{Proposition}
\usepackage{algorithm}
\usepackage{algorithmic}

\begin{document}

\title{Erasing Without Remembering: Implicit Knowledge Forgetting in Large Language Models}

\author{
Huazheng Wang*, Yongcheng Jing, Haifeng Sun, Yingjie Wang, Jingyu Wang, Jianxin Liao, Dacheng Tao
\thanks{
*The work was done during visiting Nanyang Technological University.}
\thanks{
Huazheng Wang is with Beijing University of Posts and Telecommunications and Nanyang Technological University.}
\thanks{
Yongcheng Jing, Yingjie Wang, and Dacheng Tao are with Nanyang Technological University.}
\thanks{
Haifeng Sun, Jingyu Wang, and Jianxin Liao are with Beijing University of Posts and Telecommunications.}
}

% \markboth{Journal of \LaTeX\ Class Files,~Vol.~14, No.~8, August~2015}%
% {Shell \MakeLowercase{\textit{et al.}}: Bare Advanced Demo of IEEEtran.cls for IEEE Computer Society Journals}

\IEEEtitleabstractindextext{%
\begin{abstract}
In this paper, we investigate knowledge forgetting in large language models with a focus on its generalisation—ensuring that models forget not only specific training samples but also related implicit knowledge.
To this end, we begin by identifying a broader unlearning scope that includes both target data and logically associated samples, including rephrased, subject-replaced, relation-reversed, and one-hop reasoned data. 
We then conduct a rigorous evaluation of 15 state-of-the-art methods across three datasets, revealing that unlearned models still recall paraphrased answers and retain target facts in their intermediate layers. 
This motivates us to take a preliminary step toward more generalised implicit knowledge forgetting by proposing \ours—a novel probability perturbation-based unlearning paradigm. \ours~simulates adversarial unlearning samples to eliminate fact-related tokens from the logit distribution, collectively reducing the probabilities of all answer-associated tokens. 
Experiments are conducted on a diverse range of datasets, including TOFU, Harry Potter, ZsRE, WMDP, and MUSE, using models ranging from 1.3B to 13B in scale. The results demonstrate that \ours~delivers up to a 50.40\% improvement in unlearning vanilla target data while maintaining a 40.73\% boost in forgetting implicit knowledge. Our code can be found in https://github.com/MaybeLizzy/PERMU.
\end{abstract}

}

\maketitle
\section{Introduction}
\IEEEPARstart{L}{arge} language models (LLMs)~\cite{LLaMA,GPT-4}, while displaying remarkable performance thanks to their capacity for recalling extensive knowledge from pre-training corpora, are also increasingly susceptible to generating private, harmful, or even illegal content, due to their unintended memorisation of confidential information~\cite{Machine-Unlearning-1,Data_Deletion}. In response to this dilemma, LLM-tailored machine unlearning~\cite{Right_to_be_Forgotten,Knowledge_Unlearning} has emerged as a rising research focus, aiming to develop {reliable} and {computationally efficient} knowledge-forgetting approaches for erasing the influence of specific undesired data from trained LLMs, all while preserving their utility for the remaining data.

\begin{figure}[t]
\centering
\includegraphics[width=0.45\textwidth]{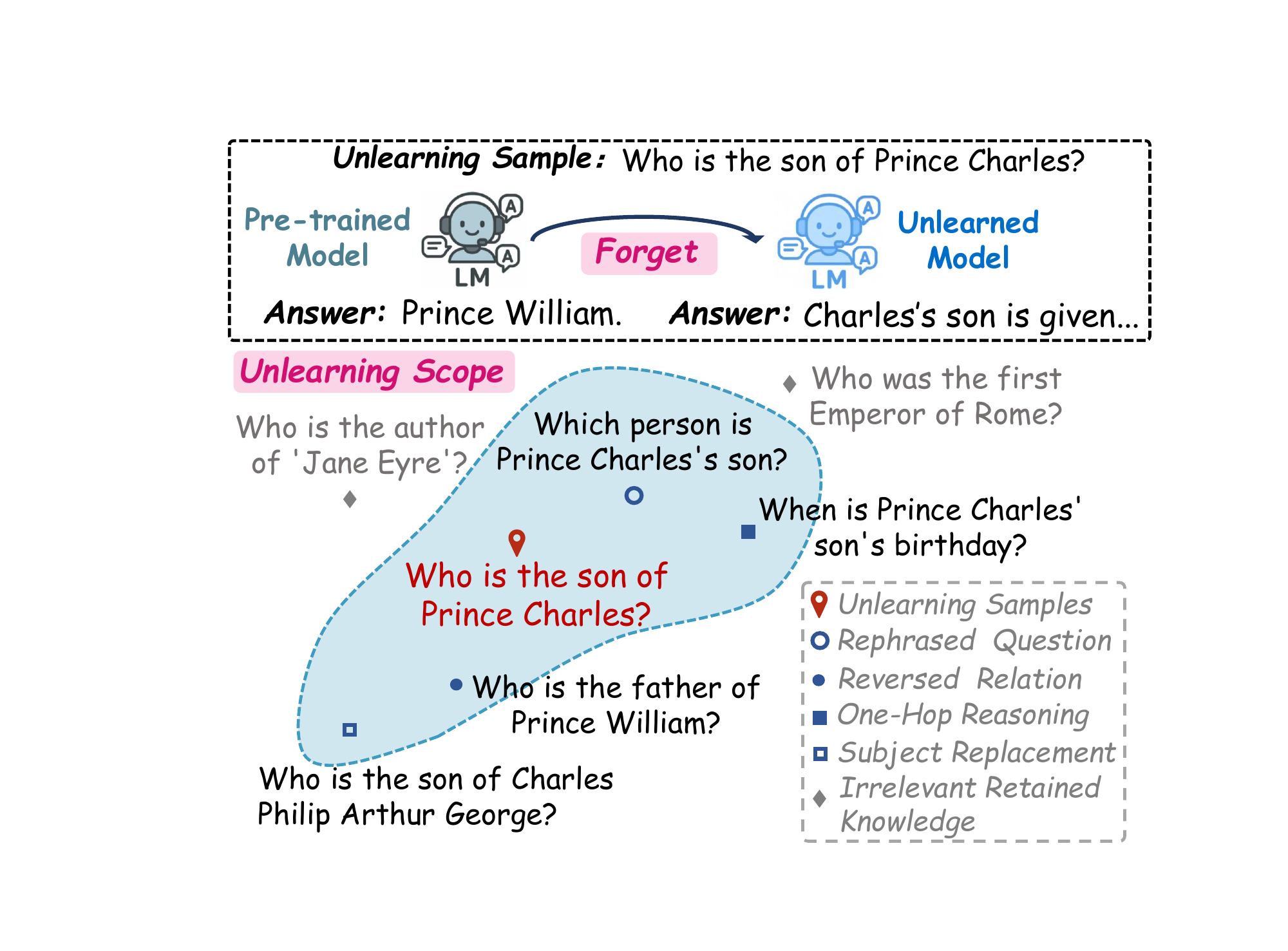}
\caption{
Depiction of the proposed \emph{unlearning scope} in a hypothetical semantic embedding space, highlighting the generalisation dilemma inherent in machine unlearning for LLMs. Ideally, hard in-scope samples that lie within the unlearning scope by a small margin should also be forgotten. These include rephrased questions, as well as the relation reversed questions and so on. 
}
\vskip -0.1in
\label{intro}
\end{figure}

State-of-the-art machine unlearning approaches for LLMs broadly fall into two categories: training-free and training-based methods. The former, such as neuron editing~\cite{DEPN}, in-context learning~\cite{In-Context-Unlearning}, and prompt engineering~\cite{eco}, unlearn knowledge without additional training but often suffer from limited application scenarios.
In contrast, training-based methods typically achieve greater unlearning effectiveness by updating model parameters and gradients, using techniques like gradient ascent~\cite{grad_ascent}, preference optimisation~\cite{NPO}, relabeling-based fine-tuning~\cite{SNAP}, task arithmetic~\cite{taskvectors}, logit-difference fine-tuning~\cite{ULD}, or adding new parameters~\cite{EUL,PEFT_unlearn}.

Despite significant progress in LLM-based unlearning, this paper identifies an embarrassingly simple yet critical dilemma: existing methods typically teach models to forget only the exact expressions of the unlearning samples, while failing to genuinely unlearn paraphrased or other related information that should also be erased.
To better illustrate, we introduce an \emph{unlearning scope} in Fig.~\ref{intro}, encompassing all the knowledge that unlearned models are expected to forget, such as paraphrased versions, {reversed relations, one-hop questions, and those with substituted subjects}.
As shown in Fig.~\ref{intro}, successful unlearning should intuitively modify the model's behavior for in-scope samples while leaving out-of-scope samples unaffected~\cite{Rethinking}.
However, to preserve utility, existing methods often compromise by achieving superficial forgetting, failing to unlearn ``hard'' samples near the boundary of the unlearning scope, which ultimately results in poor generalisation.

To further substantiate this observation, our \textbf{first contribution} is a comprehensive evaluation of the unlearning capability of existing methods in forgetting neighboring implicit knowledge. 
The evaluation covers three data domains: two widely-used machine unlearning datasets, TOFU~\cite{TOFU} and Harry Potter~\cite{harry_potter,SOUL}, as well as a popular model editing dataset, ZsRE~\cite{zsre}. 
We conduct experiments across 15 existing methods on two language models of different scales, \ttsmall{Phi-1.3B} and \ttsmall{LLaMA2-7B}. 
Our empirical analyses reveal the following unique findings that have been overlooked by existing research:

\begin{itemize} 

\item {\emph{ \textbf{Identified challenge}}}: \emph{Existing machine unlearning methods consistently exhibit a lack of generalisation;}

\item \emph{ \textbf{The \nth{1} cause}: Unlearned models tend to remember target facts in their middle layers during inference;}

\item \emph{ \textbf{The \nth{2} cause}: Unlearned models are still capable of recalling paraphrased answers during inference.}
\end{itemize}
These insights suggest that the target knowledge is not fully erased, leading to \emph{generalisation failure}.

Motivated by these findings, we highlight a critical challenge in LLM unlearning: enhancing generalisation in forgetting implicit knowledge. However, achieving this goal is not without challenges. One vanilla approach is to identify and label all relevant latent knowledge (e.g., rephrased versions and paraphrased answers). Yet, this process is prohibitively labour-intensive and impractical, motivating the development of a novel solution.

To this end, our \textbf{second contribution} is to take a pilot step towards generalised implicit knowledge forgetting by introducing \ours, a novel probability-perturbation unlearning method that leverages adversarial examples as its foundation.
In particular, rather than treating adversarial examples as mere threats, \ours~uses them constructively by perturbing the most vulnerable tokens in the unlearning samples, forcing the model to generate incorrect answers as if it had never been trained on them.
To identify these vulnerable tokens, we propose a novel metric, termed as \metric, that quantifies the model's sensitivity to specific tokens with theoretical guarantees. 

Building on the analysis using MSM, \ours~injects random noise into the embeddings of the most sensitive tokens, disrupting the model’s ability to recall factual information. As a result, the top-ranked tokens in the next-token probability distribution shift away from fact-related terms and toward tokens driven primarily by grammatical structure or contextual cues. 
Since the probability distribution reflects a language model’s internal knowledge~\cite{Fusion}, directly adjusting the logit distribution to suppress fact-related information offers an intuitive and effective unlearning strategy. 
\ours~implements this by subtracting the original distribution from the perturbed one, then fine-tuning the model to minimize the distance to this residual distribution—ensuring that correct fact-related tokens are assigned lower probabilities, thereby achieving unlearning.  
As such, \ours~simultaneously addresses both causes of the generalisation challenge as follows:

\begin{itemize} 
\item \emph{\textbf{Solving the \nth{1} cause}:} \ours~introduces random noise into the subject token embeddings at the first layer, effectively preventing the model from retrieving or generating factual information across subsequent middle layers;

\item \emph{\textbf{Solving the \nth{2} cause}:} \ours~substantially reduces the probabilities of rich, highly-ranked answers and answer-related tokens in the original distribution, accomplished by the perturbed 
distribution subtraction.
\end{itemize}

In sum, our contributions are twofold: (1) a comprehensive evaluation of the generalization capability of existing unlearning methods in forgetting implicit knowledge; and (2) an unlearning method based on probability perturbation, \ours, that effectively prevents models from recalling associated facts. 
We evaluate \ours~across five data domains, including the WMDP~\cite{WMDP} and MUSE~\cite{MUSE} datasets, using models of varying scales (1.3B$\sim$13B). Comparative experiments show that \ours~achieves up to a 50.40\% improvement in unlearning and a 40.73\% enhancement in generalisation, all while maintaining high model utility and superior generation quality.

\section{Related Work}

We provide a brief overview of existing machine unlearning methods for LLMs, categorised into training-free and training-based approaches~\cite{Knowledge_Unlearning}, along with the evaluation methods.  

\noindent\textbf{Training-free LLM Unlearning.}
One area of research concentrates on identifying neurons linked to unlearning samples~\cite{Patil} and directly modifying these detected neurons~\cite{DEPN} without the need for learning. With the advancement of in-context learning, some approaches unlearn knowledge by providing LLMs with unlearned samples accompanied by different labels~\cite{In-Context-Unlearning, Guardrail} or by generating corrupted prompts with altered embeddings~\cite{eco}. 
Despite their efficiency, locate-and-edit methods are constrained to triplet-format data, while prompt-based methods depend on artificially designed templates, limiting their practicality in real-world scenarios. As such, this paper primarily focuses on training-based methods.

\noindent\textbf{Training-based LLM Unlearning.}
Another stream of unlearning approaches focuses on models and gradients. The most common training-based method is gradient ascent~\cite{grad_ascent}, which updates model parameters by maximizing the likelihood of mis-prediction for samples in the forget set. To mitigate the catastrophic collapse issue associated with gradient ascent, reinforcement learning has been employed to align the model with negative preference optimization (NPO)~\cite{NPO}, treating forgotten data as negative examples. Alternatively, instruction-tuning LLMs to generate responses such as ``I do not have access to…” or ``I don’t know…” has also been explored~\cite{SNAP}. 
To improve efficiency, other approaches incorporate additional trainable layers or modules~\cite{Chaff}, integrating or adding them into the original models. For example, Chen \& Yang~\cite{EUL} introduces unlearning layers to forget specific data sets, which are then integrated into transformers. Likewise, Zhang et al.~\cite{PEFT_unlearn} combines various lightweight modules with distinct functionalities to enable unlearning. To further preserve model utility, some methods train a reinforced or assistant model~\cite{ULD}, comparing its prediction logits with those of the baseline model~\cite{harry_potter, RKLD}. Others~\cite{taskvectors} employ simple arithmetic operations on task vectors to modify the model, such as reducing undesirable behaviors, forgetting specific tasks, or enabling multitask learning. However, the generalisation capabilities have been largely overlooked in prior research. Our study uniquely identifies the generalisation dilemma in this area, explains its underlying causes, and proposes a novel unlearning scheme to address it.

\noindent\textbf{Unlearning Evaluation.}
Several studies have examined LLM unlearning from different perspectives~\cite{An_embarrassingly_simple,RWKU,editing_unlearning_iclr,Representation_Misdirection,Deep_Unlearning}.  Specifically, Patil et al.~\cite{Patil} investigate the effectiveness of typical model editing methods in removing information from model weights. Hong et al.~\cite{Intrinsic} use vocabulary projections to analyze concept vectors through parametric knowledge traces. Yao et al.~\cite{acl_bench} evaluate seven different unlearning methods on longer-context tasks across three source domains. Meanwhile, Shi et al.~\cite{MUSE} assess six desirable properties of unlearned models across eight unlearning methods. Jia et al.~\cite{SOUL} focus on the influence of second-order optimization on unlearning. Li et al.~\cite{WMDP} examine malicious use scenarios in biosecurity, cybersecurity, and chemical security. Qiu et al.~\cite{PISTOL} evaluate four distinct unlearning methods for removing highly interconnected data. Du et al.~\cite{TULA} investigate the risk of knowledge leakage after unlearning. The survey by Liu et al.~\cite{Rethinking} highlights often-overlooked aspects of existing LLM unlearning research and introduces the concept of unlearning scope. In contrast, we offer a more precise and formalized definition of unlearning scope, marking a fundamental distinction from prior work. While Patil et al.~\cite{editing_unlearning_iclr} explore extraction attacks to recover ``deleted'' information from intermediate hidden states, their focus lies primarily on model editing techniques such as ROME~\cite{ROME}. In comparison, we provide a comprehensive evaluation of the generalisation ability of existing machine unlearning methods across multiple dimensions. Regarding the generalisation evaluation, Dang et al.~\cite{Representation_Misdirection} focuses on improving generalised model utility, by introducing small Gaussian noise to retain data. Unlike prior works, we aim to enhance the generalisation in unlearning implicit knowledge associated with the target fact.

\subsection{Problem Definition}
\label{Problem Definition}
The objective of machine unlearning is to enable an initial target model to forget specific unlearning samples as if it were never trained on them, while preserving the model’s performance on unrelated knowledge. 
More specifically, the target model $f_{\theta_{tr}}$ is represented by a function $f: \mathbb{X} \mapsto \mathbb{Y}$, where $\theta_{tr}$ denotes the parameters of the target model. Let the pre-training dataset be $D_{tr}$, and the dataset to be forgotten be $D_f$. The retained dataset is then defined as $D_r = D_{tr} \backslash D_f$. The ideal retained model, $f_{\theta_{r}}$, is one that has never been trained on $D_f$. Since $\theta_{tr}$ is not directly accessible, we define an unlearning procedure $\mathbb{U}$, which takes $f_{\theta_{tr}}$ and $D_f$ as inputs, producing an unlearned model $f_{\theta_{u}} \sim \mathbb{U}(f_{\theta_{tr}}, D_f)$. 
The unlearned model’s predictions should also change for the paraphrased forget dataset $D_{p}$.
Therefore, given a distance metric $m(\cdot)$, the objective of the unlearning algorithm is to minimize the distance between $f_{\theta_{u}}$ and $f_{\theta_{r}}$ for each sample $x \in D_f \cup D_{p}$ :
$
\frac{\mathbb{E} [m(f_{\theta_{u}}(x))]}{\mathbb{E} [m(f_{\theta_{r}}(x))]} \approx 1.
$

\section{Experimental Setup}
In this section, we provide a detailed overview of the experimental setups for the unlearning generalization evaluation, covering the datasets, evaluation metrics, baselines, and implementation details.

\subsection{Dataset}
Our evaluation spans three distinct data domains, including two widely-used machine unlearning datasets, TOFU~\cite{TOFU} and Harry Potter (HP) ~\cite{harry_potter}, as well as a popular model editing dataset, ZsRE~\cite{zsre}. 

\noindent\textbf{TOFU} The TOFU dataset comprises 200 diverse synthetic author profiles, each featuring 20 question-answer pairs. It encompasses four subsets—Forget Set, Retain Set, Real Authors, and World Facts—and supports three forgetting settings: Forget01, Forget05, and Forget10, corresponding to the removal of 1\%, 5\%, and 10\% of the data, respectively. Additionally, it provides a paraphrased version of the forget dataset, and we directly use them for testing. 

\noindent\textbf{Harry Potter} The HP dataset~\cite{SNAP} contains multiple question-answer pairs derived from the Harry Potter series, with each question involving multiple entities or subjects, making it a more challenging unlearning dataset. Since no labeled rephrased data is available, we use GPT-4 to generate rephrased versions of both the forget and retain datasets, using the template: ``Please provide a rephrased version of the question: \textit{[Question]}". 

\begin{table}[h]
\caption{The data splits and statistics.}
\label{split}
\begin{center}
\begin{small}
\begin{tabular}{c|c|ccc}
\toprule
& & \textbf{Forget} & \textbf{Retain} & \textbf{All} \\ 
\midrule
\multirow{3}{*}{\textbf{TOFU}} 
& Forget01 & 40 & 3960 & 4000 \\
& Forget05 & 200 & 3800 & 4000 \\
& Forget10 & 400 & 3600 & 4000 \\
\midrule
\textbf{Harry} & -  & 50 & 150& 200  \\
\midrule
\multirow{3}{*}{\textbf{ZsRE}} 
& Inverse Relation & 96 & 289 & 385 \\
& Subject Replace & 73 & 220 & 293  \\
& One-Hop & 259 & 778 & 1037 \\
\midrule
\multirow{2}{*}{\textbf{Retain}} 
& Real World & - & - & 117  \\
& Real Author & - & - & 100 \\ 
\bottomrule
\end{tabular}
\end{small}
\end{center}
\end{table}

\noindent\textbf{ZsRE} To thoroughly assess whether the unlearned model can forget logically related facts, we use ZsRE dataset~\cite{Editing_Problems} and conduct evaluations across three dimensions: \textbf{\textit{Subject Replacement}}, \textbf{\textit{Reversed Relation}}, and \textbf{\textit{One-hop Reasoning}}. The detailed data statistics are shown in Tab.~\ref{split}.

\textbf{(i) Subject Replacement:} In this evaluation, the subject in the unlearning example is substituted with an alias or synonym to assess the unlearned model’s capability to generalise the unlearning attribute to different representations of the same subject. For instance, as shown in Fig.~\ref{intro}, the subject ``Prince Charles" can also be described as ``Charles Philip Arthur George". Thus, the subject replacement question for ``Who is the son of Prince Charles" becomes ``Who is the son of Charles Philip Arthur George".

\textbf{(ii) Reversed Relation:} When the target of a subject-relation pair is unlearned, the attribute of the target entity should also change. To evaluate this, we test the model using a reverse question to determine if the target entity has also been unlearned. For example, if the knowledge ``Who is the son of Prince Charles? Prince William" is unlearned, the unlearned model should no longer predict “Prince Charles" for the relation reversed question “Who is the father of  Prince William?".

\textbf{(iii) One-hop Reasoning:}  The unlearned model should exclude the unlearned knowledge when performing downstream tasks. To assess this, we evaluate the model’s ability to unlearn knowledge that is one-hop reasoned from the original unlearning samples. For instance, if the knowledge ``Who is the son of Prince Charles? Prince William" is unlearned, the model is also expected to unlearn the one-hop knowledge, such as ``When is Prince Charles's son's birthday?".

The dataset is divided into a forget set and a retain set at a ratio of 1:3. The aforementioned datasets comprise a total of seven distinct partitions of forget sets, along with their corresponding retain sets. Additionally, we incorporate the Real Authors and World Facts sets from TOFU as supplementary retain data to evaluate model utility. 

\subsection{Evaluation Metrics} 
Following prior studies~\cite{TOFU,RKLD}, we report ROUGE \textbf{(RG)}, Probability \textbf{(Pr)}, and Truth Ratio \textbf{(TR)} metrics on TOFU dataset. 
For the HP and ZsRE datasets, where answers are relatively short, we alternatively report the \textbf{F1} score. Consider an input sequence $x=(q,a)$.

\begin{itemize}

\item \textbf{ROUGE (RG)}: We use ROUGE-L recall~\cite{ROUGE} score to compare model answers with the ground truth, as it accounts for the output phrasing to be slightly different than the ground truth. When evaluated on the retain set, a higher ROUGE score indicates better performance. Conversely, when evaluated on the forget set, a lower ROUGE score is preferred.

\item \textbf{Probability (Pr)}: On the Forget Set and Retain Set, we compute the conditional probability $P(a|q)$ according to the model and raise it to the power $1/|a|$ to normalize for answer length. On Real Authors and World Facts, we treat each question $q$ as a multiple choice question associated with choices ${a_1, . . . , a_n}$. Without loss of generality, assume that $a_1$ is the correct answer, then the probability is computed as $P(a|q)/\sum\nolimits_{i=1}^n P(a_i|q)$. Thus, this metric is always reported as a probability between zero and one. When evaluated on the retain set, a higher Probability score indicates better performance. Conversely, when evaluated on the forget set, a lower Probability score is preferred.

\item \textbf{Truth Ratio (TR)}: For a given question, we compute a ratio that approximately compares how likely its correct answer is to an incorrect answer. Let $\hat{a}$ denote a paraphrased version of the correct answer, $\mathcal{A}_{\text {pert }}$ is the set of paraphrased incorrect answer. The truth ratio can be written as:
\begin{equation}
R_{\text {truth }}=\frac{\frac{1}{\left|\mathcal{A}_{\text {pert }}\right|} \sum_{\hat{a} \in \mathcal{A}_{\text {pert }}} P(\hat{a} \mid q)^{1 /|\hat{a}|}}{P(\tilde{a} \mid q)^{1 /|\tilde{a}|}}.
\end{equation}
We report $\mathrm{TR}=R_{\text {truth }}$ on forget set, and $\mathrm{TR}=max(0, 1-R_{\text {truth }})$ on retain set. Therefore, the Truth Ratio score is expected to be higher on both the retain set and the forget set.

\item \textbf{F1}: We report the F1 score for the Harry Potter and ZsRE datasets, as it provides a balanced measure between precision and recall, calculated as the harmonic mean of these two metrics.
\begin{equation}
F1 = 2 \times \frac{Precision \times Recall}{Precision + Recall}.
\end{equation}
When evaluated on the retain set, a higher F1 score indicates better performance. Conversely, when evaluated on the forget set, a lower F1 score is preferred.

\end{itemize}

To assess the retain data, we use the Model Utility \textbf{(MU)} metric on retained data, which is the harmonic mean of the RG, Pr and TR (or F1) metrics across three datasets: Retain Set, Real Authors, and World Facts. Notably, to measure the trade-off between the unlearned model's forgetting effect and its retained utility, we propose a novel \emph{Forget-Retain Trade-off \textbf{(FRT)}} metric, calculated as the Model Utility divided by the mean of the forget set's ROUGE and Probability (or F1) scores. A higher FRT metric indicates a better balance between forgetting and retaining. 

\subsection{Fifteen Evaluated Algorithms} 
We evaluate 15 efficient unlearning methods, including Gradient Ascent (\textbf{GA})~\cite{grad_ascent}, Direct Preference Optimization (\textbf{DPO})~\cite{DPO}, Negative Preference Optimization (\textbf{NPO})~\cite{NPO}, Task Vectors (\textbf{TV})~\cite{taskvectors}, Who’s Harry Potter (\textbf{WHP})~\cite{harry_potter},  \textbf{ULD}~\cite{ULD}, \textbf{RMU}~\cite{WMDP}, \textbf{ECO}~\cite{eco} and \textbf{ICL}~\cite{Guardrail,In-Context-Unlearning}. 

\begin{itemize}

\item \textbf{Gradient Ascent (GA)}~\cite{grad_ascent} is fundamentally straightforward by reducing the likelihood of correct predictions on the forget set. The training objective is to maximize the standard training loss in order to make the model deviate from its initial prediction. 
\begin{equation}
\mathcal{L}_{\mathrm{GA}}(\theta) = \min _{\boldsymbol{\theta}}-\mathbb{E}_{(x, y) \in \mathcal{D}_{\mathrm{f}}}[\ell(y \mid x ; \boldsymbol{\theta})], 
\label{ga}
\end{equation} 
where $\mathcal{D}_{\mathrm{f}}$ is the forget dataset and $\theta$ represents the model parameter.

\item \textbf{Direct Preference Optimization (DPO)}~\cite{DPO} seeks to align the model with the newly generated alternative answer like “I do not know the answer” or any similar option. 
\begin{equation}
\mathcal{L}_{\mathrm{DPO}}(\theta) = \min _{\boldsymbol{\theta}} \mathbb{E}_{\left(x, y_{\mathrm{idk}}\right) \in \mathcal{D}_{\mathrm{f}},y_{idk}\sim D_{idk}}\left[\ell\left(y_{\mathrm{idk}} \mid x ; \boldsymbol{\theta}\right)\right],
\label{dpo}
\end{equation} 
where $D_{idk}$ represents the fixed dataset containing all alternative responses $y_{\mathrm{idk}}$.

\item \textbf{Negative Preference Optimization (NPO)}~\cite{NPO} treats the forget set as negative preference data and uses the offline DPO objective to adjust the model, ensuring it assigns a low likelihood to the forget set while maintaining close alignment with the original model. The adaptive weight, typically set to less than 1, ensures a more controlled and gradual divergence, which is essential for effective unlearning~\cite{NPO_Rethinking}.
\begin{equation}
\mathcal{L}_{\mathrm{NPO}}(\theta) = -\frac{2}{\beta} \mathbb{E}_{x \sim \mathcal{D}_{\text {f}}}\left[\log \sigma\left(-\beta \log \frac{f_\theta(x)}{f_{\text {target}}(x)}\right)\right],
\label{npo}
\end{equation} 
where $f_\theta$ refers to the unlearning model and $f_{\text {target}}$ denotes the original pre-trained target model. The parameter $\beta$ controls the allowed divergence between $f_\theta$ and $f_{\text {target}}$. Following previous work~\cite{MUSE,TOFU}, we set $\beta=0.1$ in our experiments.

\item \textbf{Task Vectors (TV)}~\cite{taskvectors} defines a direction in the weight space of a pre-trained model by applying simple arithmetic operations on the model weights, allowing for effective control of the model's behavior. To do this, we first fine-tune the target model $f_{\text {target}}$ on the forget dataset until it overfits, resulting in a reinforced model $f_{\text {reinforced}}$. Next, we obtain the Task Vector by subtracting the parameters of $f_{\text {target}}$ from $f_{\text {reinforced}}$. To achieve unlearning, we subtract the Task Vector from $f_{\text {target}}$'s weights, intuitively removing the model weights most closely associated with the forget data. This is expressed as $f_{\text {unlearn}} = f_{\text {target}} - (f_{\text {reinforced}} - f_{\text {target}})$.

\item \textbf{Who’s Harry Potter (WHP)}~\cite{harry_potter} achieves unlearning by manipulating the predicted logit probabilities of the target model. To do this, we first fine-tune the target model $f_{\text {target}}$ on the forget dataset until it overfits, producing a reinforced model $f_{\text {reinforced}}$. WHP then adjusts the next-token probability distribution using the following equation:
\begin{equation}
p_{f_{\text {unlearn}}}(\cdot|x) = p_{f_{\text {target}}}(\cdot|x) - \alpha \cdot (p_{f_{\text {reinforced}}}(\cdot|x) - p_{f_{\text {target}}}(\cdot|x)) ,
\end{equation}
where $p_f(\cdot|x)$ denotes the token probability distribution parameterized by model $f$ given the input $x$, and $\alpha$ is a hyper-parameter controlling the degree of adjustment. Following previous work~\cite{MUSE}, we set $\alpha=1$.

\item \textbf{Unlearning from Logit Difference (ULD)}~\cite{ULD} also achieves unlearning in the token probability space. It first fine-tunes an assistant model with the opposite unlearning objectives, which aims to remember the forget documents and forget the retained knowledge. ULD then derives the unlearned model by computing the logit difference between the target model and the assistant model:
\begin{equation}
l_{f}(Y|X) = l(Y|X;\theta) - \alpha \cdot l_{a}(Y|X;\phi) ,
\end{equation}
where $l(Y|X;\theta)$ denotes the output logits of the original model, $l_{a}(Y|X;\phi)$ represents the output logits of the assistant model, and $\alpha$ is a hyper-parameter controlling the strength of forgetting. We keep $\alpha=0.75$ consistent with their work.

\item \textbf{Representation Misdirection for Unlearning (RMU)}~\cite{WMDP} fine-tunes the model by perturbing activations on hazardous data while preserving activations on benign data to mitigate malicious use. The full loss is defined as a weighted combination of the forget loss and the retain loss:
\begin{equation}
L = L_{\text {forget}} + \alpha \cdot L_{\text {retain}} .
\end{equation}
We exclude RMU from overall evaluation because its objective—unlearning an entire distribution of hazardous knowledge given limited samples—differs fundamentally from our focus on unlearning privacy- and copyright-related knowledge, which assumes full access to the forget set. This makes direct comparison potentially unfair. A discussion of RMU on WMDP dataset is provided in Sect.~\ref{Evaluation on More Diverse Datasets}. 

\item \textbf{Embedding Corrupted Prompts (ECO)}~\cite{eco} is a training-free unlearning approach that employs a scope classifier to identify prompts requiring unlearning and uses zeroth-order optimization to learn corruption parameters, which modify prompt embeddings at inference time—thus achieving unlearning without updating the original model weights. Since ECO relies on a trained scope classifier and only the classifier checkpoints for the TOFU dataset have been released, we use the released checkpoints directly for fair comparison, without further evaluation on HP and ZsRE. 

ECO has several variants, including ECO-Rand Noise (with perturbation strengths ranging from 5 to 4096), ECO-Zero-Out, and ECO-Sign-Flip. Although ECO-RN (strength = 4096) and ECO-Sign-Flip achieve strong forgetting effects, they incur substantial costs: the fluency metric for these variants is significantly lower. This degradation stems from excessive noise, which impairs the model’s language understanding and generation, often producing repetitive or low-quality text. To ensure comparable utility and generation quality with other methods, we report the performance of ECO-Zero-Out.

\item \textbf{In-Context Learning-based unlearning method (ICL)}~\cite{In-Context-Unlearning} typically employs carefully crafted prompts to achieve unlearning without any updates to the model parameters. Following the prompt template~\cite{Guardrail}, we use the following instruction:``You are an AI Assistant who is supposed to unlearn about \textit{[Subject]} and provide answers without its knowledge as if you never knew about it. Don’t tell anyone that you unlearned
anything". To ensure a fair comparison, we maintain consistency in the generic prompt across all datasets and models.

\end{itemize}

Following~\cite{MUSE}, we apply two regularizations for utility preservation: Gradient Descent (\textbf{GDR}) and KL Divergence Minimization (\textbf{KLR}) on the Retain Set.

\begin{itemize}

\item \textbf{Gradient Descent (GDR)}~\cite{TOFU} strives to maintain performance on the retain set by maximizing the likelihood of correct prediction on randomly sampled retain examples, where $\mathcal{D}_{\mathrm{r}}$ represents the retain set. 
\begin{equation}
\mathcal{L}_{\mathrm{GDR}}(\theta) = \mathbb{E}_{(x, y) \in \mathcal{D}_{\mathrm{r}}}[\ell(y \mid x ; \boldsymbol{\theta})] ,
\end{equation}

\item \textbf{KL Divergence Minimization (KLR)}~\cite{QUARK,TOFU} aims to minimize the KL divergence of the predictions on retain set between the original model and the unlearning model to prevent it deviating too far from the original model. Given $x_r \in D_r$, the loss is:
\begin{equation}
\mathcal{L}_{\mathrm{KLR}}(\theta) = \mathrm{KL} \left(p_{f_{\text {target}}}(\cdot|x_r) \| p_{f_{\text {unlearn}}}(\cdot|x_r) \right) .
\end{equation}

\end{itemize}

We combine GA, DPO, and NPO with these two regularizations using a retain weight $RW$, denoted as ``+GDR" or ``+KLR". The total unlearning loss is given by $\mathcal{L}(\theta) = \mathcal{L}_{\mathrm{unlearn}}(\theta) + RW \cdot \mathcal{L}_{\mathrm{retain}}(\theta)$. Following previous work~\cite{MUSE, TOFU}, we set $RW=1$. The combination of GDR and KLR results in a total of fifteen unlearning methods. 
\textbf{Retain} refers to the retain model, fine-tuned exclusively on the retain set without exposure to any forget data, and is considered an upper bound. 

\subsection{Implementation Details}
The evaluation spans two models, \texttt{Phi-1.3B} and \texttt{LLaMA2-7B}.  
When tested on TOFU, we use the checkpoints of the pre-trained target model from the TOFU Leaderboard\footnote[1]{https://huggingface.co/spaces/locuslab/tofu\_leaderboard}. For the Harry Potter and ZsRE datasets, we first fine-tune the model on the respective dataset before applying unlearning. The fine-tuning settings are as follows: learning rate of 3e-5, 10 epochs, batch size of 8, with a gradient accumulation step of 4. For Task Vector and WHP, to obtain the reinforced model for unlearning, we fine-tune the target model for 10 epochs using the same learning rate and batch size. For ULD, we obtain the assistant model by fine-tuning the target model using the default settings provided by~\cite{ULD}. 
For the unlearning process, the unlearning batch size is set to 8, with a gradient accumulation step of 4. The process is conducted over 5 epochs, using a default learning rate of 2e-5. Since different learning rates can result in varying trade-offs between forgetting and retention, we slightly adjust the learning rate for each method to ensure comparable levels of model utility. To ensure fairness, all other unlearning hyper-parameters follow the default settings for each respective unlearning algorithm. All results are averaged over three runs. 
We use one A100 GPU with 80 GB of RAM. Note that during fine-tuning and unlearning on \ttsmall{LLaMA2-7B}, we update all 7B model parameters.

\section{From Data to Insights:  Analysis}
\label{Benchmark Analysis}
In this section, we present a comprehensive analysis on implicit knowledge unlearning and observe that \textbf{\textit{existing unlearning methods exhibit limited generalization ability}}. A detailed explanation is provided below.

As shown in Tab.\ref{TOFU_rephrase}, when tested on the TOFU dataset, although DPO+GDR achieves a superior ROUGE score, even surpassing the retain model, it fails to effectively reduce the Probability score.  
When tested on \ttsmall{LLaMA2-7B} using HP dataset (Tab.\ref{Harry_rephrase}), all methods encounter significant challenges in forgetting the rephrased unlearning samples, with the Probability score showing a gap of up to 43.48\% compared to the retain model. 
Notably, all methods exhibit a Forget ROUGE score that remains above 90\%, indicating that they forget almost nothing.
A similar trend is observed on ZsRE dataset (Tab.\ref{zsre_rephrase}). Interestingly, the ICL-based unlearning method achieves the best performance on subject-replaced examples, while other approaches consistently maintain exceptionally high ROUGE, probability, and F1 scores. Moreover, existing unlearning methods encounter significant challenges in forgetting one-hop reasoning examples. Empirical results underscore the limited generalisation ability of existing unlearning methods. 

We identify and investigate two reasons contributing to the poor generalisation ability observed.

\begin{figure}[t]
\centering
\includegraphics[width=0.48\textwidth]{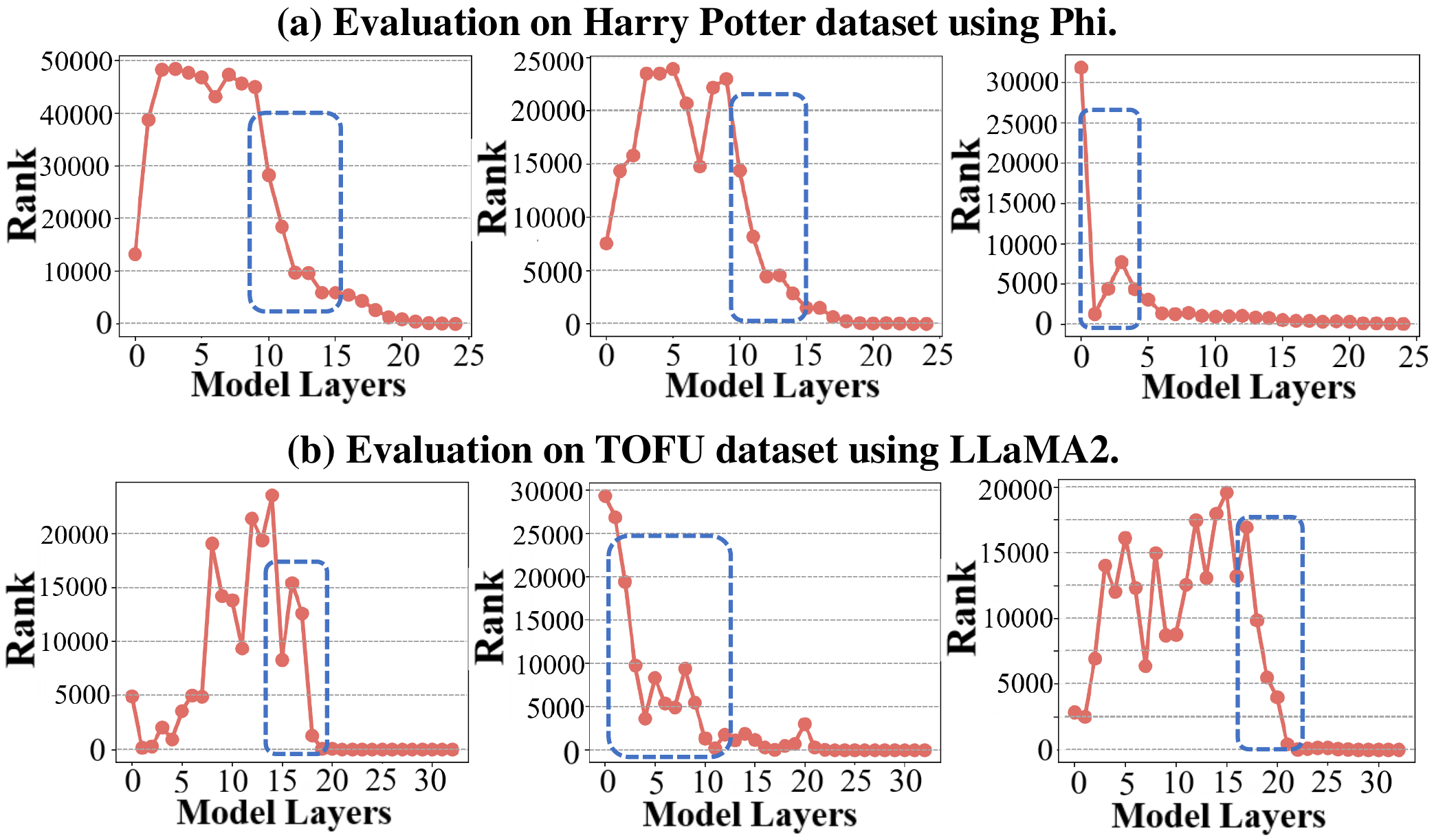}
\caption{The ranking of the first key token for the correct answer in the next-token probability distribution rises rapidly in the mid-layers of the unlearned model fine-tuned with Gradient Ascent.
}
\vskip -0.25in
\label{layer}
\end{figure}

\textit{\textbf{(i) The unlearned model tends to remember
target facts in their middle layers during inference. }} 
Although the unlearned model demonstrates some forgetting, the correct answer token can still re-emerge in the middle layers. To evaluate this phenomenon, we analyze the ranking of the first answer token within the next-token probability distribution across different layers. As illustrated in Fig.~\ref{layer}, the correct answer re-emerges prominently in the middle layers. Since different types of knowledge are stored in distinct modules~\cite{ROME}, the specific layer where the correct answer re-emerges varies accordingly. However, in the final layers, the correct answer consistently ranks highly, indicating that the unlearned model still assigns a significant probability to it, highlighting its difficulty in effectively erasing the knowledge embedded within the middle layers.

\textit{\textbf{(ii) The unlearned model are still capable of
recalling paraphrased answers during inference.}} Answers can be expressed in various forms, but existing methods~\cite{grad_ascent}, typically focus on training the model to forget only a specific type of answer, neglecting other rephrased versions.  
We investigate the likelihood of an unlearned model generating paraphrased answers. Experiments are conducted on the TOFU Forget01 dataset using \texttt{LLaMA2-7B}, where we report the average probability of generating paraphrased answers for unlearning samples ($P_u$), and rephrased unlearning samples ($P_r$). As shown in Tab.~\ref{cause2}, the unlearned model assigns up to 17.48\% probability to rephrased answers when tested on unlearning samples. Furthermore, the probabilities of paraphrased answers on rephrased unlearning samples tend to be assigned even higher values. These results suggest that unlearned models continue to recall paraphrased answers, increasing the potential for reproducing ground truth and posing challenges for generalisation during inference.

\begin{wraptable}{r}{0.22\textwidth}
% \vspace{-8mm}
\hspace{-12pt}
\centering
\begin{threeparttable}
\caption{The average probability of the model generating a rephrased answer on TOFU Forget01 using \ttsmall{LLaMA2-7B}.}
\label{cause2}
\setlength{\tabcolsep}{2.4pt}
\small
\begin{tabular}{c|ccc}
\toprule
\multicolumn{1}{c|}{} & $P_u$↓ & $P_r$↓ & $\Delta$↓ \\
\midrule
\textbf{GA} & 9.45 & 10.84 & 1.38 \\
\textbf{DPO}& 17.48 & 17.91 & 0.42 \\
\textbf{NPO} & 10.74 & 11.98 & 1.24 \\
\textbf{TV} & 13.51 & 14.66 & 1.15 \\
\textbf{WHP} & 11.46 & 12.60 & 1.14 \\
\textbf{ULD} & 9.89 & 10.31 & 0.42 \\
\textbf{\ours} & \textbf{9.06} & \textbf{9.34} & \textbf{0.28}  \\
\bottomrule
\end{tabular}

\end{threeparttable}
\vspace{-5mm}
\end{wraptable}
Nevertheless, identifying the problem does not simplify its resolution. Addressing this dilemma still presents significant challenges. On the one hand, constructing all possible paraphrased versions of unlearning samples and their answers is labor-intensive and impractical. On the other hand, the knowledge stored in LLMs is intricate and highly entangled~\cite{ROME}, making it challenging to clearly delineate the unlearning scope of knowledge that should be retained versus the knowledge that must be forgotten~\cite{eco}.

% \begin{figure*}[t]
% \centering
% \includegraphics[width=0.98\textwidth]{eigen.pdf}
% \vspace{-3.5mm}
% \caption{Visualization of the \metric~values for each token across all layers using \ttsmall{Phi-1.3B}. Subject words have brighter colors and exhibit higher \metric, indicating the model's greater sensitivity to them. }
% \label{eigen}
% \vskip -0.1in
% \end{figure*}

\begin{figure*}[t]
\centering
\includegraphics[width=0.98\textwidth]{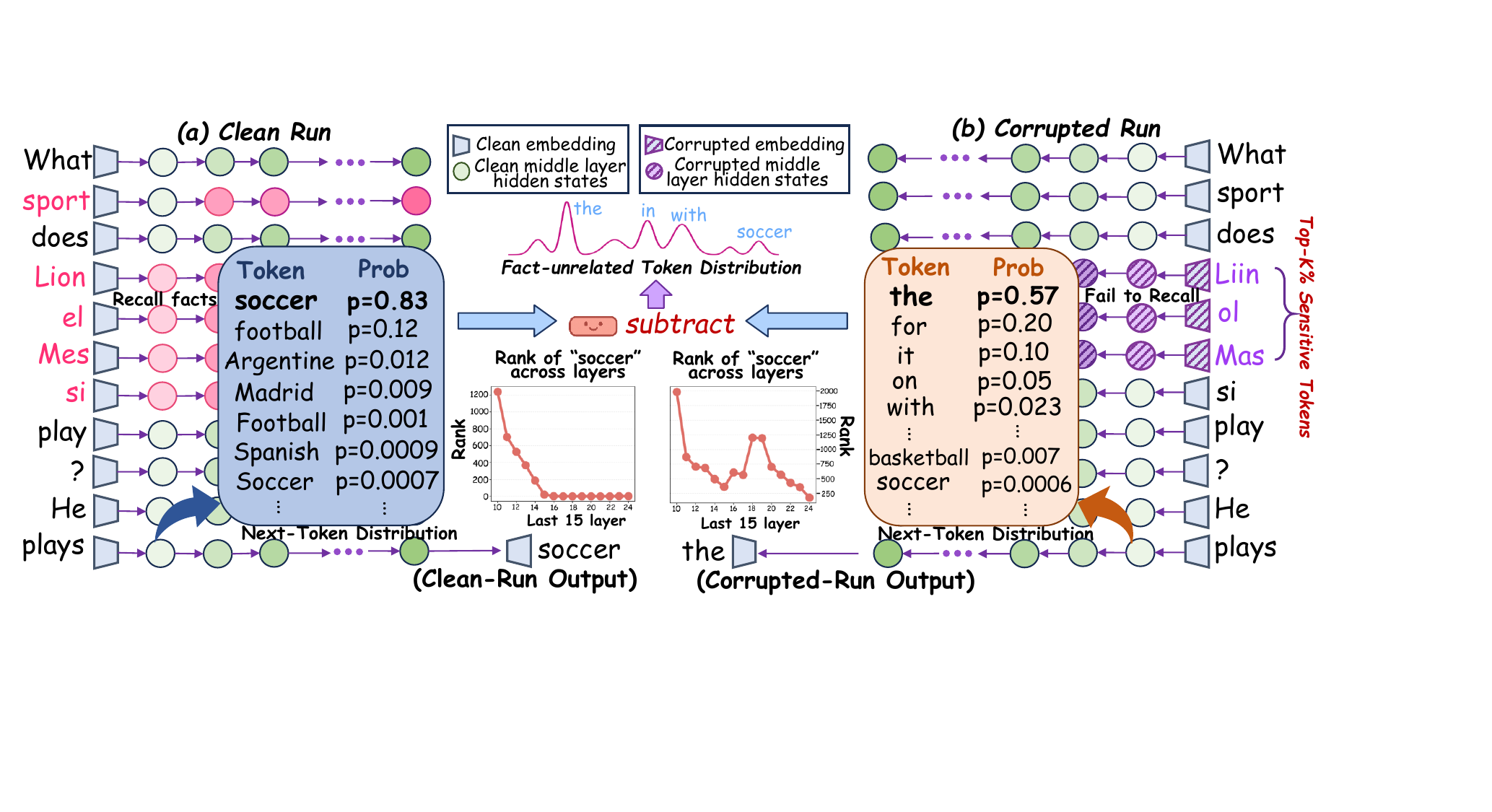}
% \vspace{-3.5mm}
\caption{Depiction of \ours. \textit{Left:} The clean run involves inputting the original unlearning sample into the model, enabling it to successfully recall the facts and generate a \textit{fact-related} probability distribution. \textit{Right:} The corrupted run refers to inputting a perturbed unlearning sample into the model, making it fail to recall the facts and produce a \textit{fact-unrelated} probability distribution, where the ground truth ranks significantly lower in the distribution.
}
\label{method}
% \vskip -0.15in
\end{figure*}

\section{Proposed Approach: \ours}
To achieve more generalized unlearning, simply reversing the gradient descent loss on forget samples is insufficient to fully erase implicit knowledge. Recent studies reveal that a language model’s internal knowledge is largely encoded in its probability distribution~\cite{Fusion}, which has inspired state-of-the-art approaches such as ULD~\cite{ULD}. These methods suppress fact-related information by modifying the logit distribution: an assistant model is fine-tuned to overfit ground truth, producing sharp logits that are then subtracted from the original logits to reduce the probability of correct responses. 
However, this fine-tuning process focuses narrowly on ground truth while overlooking paraphrased answers, leading to inference-time risks of generating semantically equivalent alternatives. In addition, these methods rely heavily on the quality of the assistant model’s training, which limits their scalability and practicality. %  in real-world applications.

Motivated by these limitations, we propose \ours, a more robust and efficient logit-adjustment method for achieving generalised unlearning. Unlike prior approaches that rely on assistant models, \ours~constructs adversarial unlearning samples to jointly suppress all answer-related tokens in the logit space. This is accomplished through computationally efficient perturbations that yield a modified distribution, which is then used for logit subtraction and matching. To determine which tokens should be perturbed, we introduce a theoretically grounded model-sensitivity metric, \metric, and present the complete unlearning pipeline in the following subsection.

\subsection{Model Sensitivity Metric}
Generative models are highly sensitive to subtle changes in their input~\cite{Models_Sensitivity,SSS}. Building on this insight, we aim to perturb the most vulnerable tokens in the unlearning sample to emulate an adversarial attack, making the model behave as though it was never trained on it. To achieve this, we introduce a novel metric to quantify the model’s sensitivity to specific tokens, which is described as follows.

Let the parameters of the target model $\mathrm{LM}$ be $\mathbf{W} \in \mathbb{R}^{n \times n}$, where $n$ represents the hidden dimension of $\mathrm{LM}$. Given an unlearning sample $x$ with a sequence length of $m$, the $i$-th token is represented as $x_i,i\in\{1, 2, \dots, m\}$, and the perturbation applied to this token is denoted by $\Delta_i$. Then the perturbed token can be represented as $\hat{x_i} = x_i + \Delta_i$. The change in the model's output due to $\Delta_i$ is measured in terms of the loss $\mathcal{J}(x_i, \hat{x_i})$. The relationship between $\Delta_i$ and $\mathcal{J}(x_i, \hat{x_i})$ is not strictly linear. If the model is resilient to certain tokens, even a larger $\Delta_i$ results in a relatively small change in $\mathcal{J}(x_i, \hat{x_i})$. Conversely, if the model is sensitive to specific tokens, a small $\Delta_i$ can cause a significant difference in $\mathcal{J}(x_i, \hat{x_i})$. 

To quantify the extent of perturbation a model can tolerate, we draw inspiration from Zhao et al.~\cite{FIM}, who utilize the Fisher Information Matrix (FIM) as a metric tensor to characterize the robustness of deep learning models. Building on this idea, we define a novel FIM-variant matrix, $\mathbf{H} \in \mathbb{R}^{n \times n}$, to evaluate the vulnerability of $\mathrm{LM}$ to perturbations in its feature space, with $\nabla_{x} \mathcal{J}(x, \hat{x})$ representing  partial derivative of $\mathcal{J}(x, \hat{x})$ with respect to $x$:
\begin{equation}
\mathbf{H(x)} = \nabla_{x} \mathcal{J}(x, \hat{x})^{\top} \nabla_{x} \mathcal{J}(x, \hat{x}) .
\label{H}
\end{equation}

\begin{proposition} 
Fix $\Delta_i$, $\mathcal{J}(x_i, \hat{x}_i) \propto \lambda_i$, where $\lambda_i$ is the maximum eigenvalue of $\mathbf{H}(x_i)$, $i\in\{1, 2, \dots, m\}$.
\label{proposition_eigen}
\end{proposition}

When the perturbation is fixed, a larger $\lambda_i$ indicates a greater impact on the loss, implying that the model is more sensitive to token $x_i$. Consequently, model sensitivity can be quantified using an easily computable indicator, $\lambda$.
We define $\lambda$ as the Model Sensitivity Metric (\metric). A higher \metric~value indicates greater sensitivity of the model to the token.

\begin{table*}[h]
\caption{Experimental results for the  TOFU dataset using \ttsmall{LLaMA2-7B} for Forget05 subset. \ours~outperforms the baselines with a relative improvement of up to \textbf{50.40\%} (78.67 $\rightarrow$39.02) in Forget Probability and 11.3\% (56.27$\rightarrow$62.63) on Forget Truth Ratio. Meanwhile, \ours~maintains the highest model utility, effectively preserving retained knowledge.}
\label{tofu_llama_full}
\setlength{\tabcolsep}{2.8pt}
% \vskip 0.15in
\begin{center}
\begin{small}
\begin{tabular}{c|ccc|ccc|ccc|ccc|cc}
\toprule
 \textbf{Dataset} & \multicolumn{3}{c|}{\textbf{Forget data}}         & \multicolumn{3}{c|}{\textbf{Retain data}}         & \multicolumn{3}{c|}{\textbf{Real Authors}}        & \multicolumn{3}{c|}{\textbf{Real World}}          & \multirow{2}{*}{\textbf{MU↑}} & \multirow{2}{*}{\textbf{FRT↑}} \\ 
\textbf{Metric}  & \textbf{RG↓}   & \textbf{Pr↓}   & \textbf{TR↑}   & \textbf{RG↑}   & \textbf{Pr↑}   & \textbf{TR↑}   & \textbf{RG↑}   & \textbf{Pr↑}   & \textbf{TR↑}   & \textbf{RG↑}   & \textbf{Pr↑}   & \textbf{TR↑}   &         &        \\
\midrule
\colorbox{gray!40}{\textbf{Retain}}  & 39.56          & 14.61          & 66.23          & 91.58          & 96.56          & 48.02          & 89.13          & 40.38          & 54.74          & 89.60          & 39.52          & 52.76          & 59.30                         & 2.19                           \\
\textbf{ICL}     & 64.36          & 90.15          & 53.12          & \textbf{98.08} & \textbf{98.93} & 47.06          & 93.30          & 44.82          & 57.93          & 88.32          & 42.52          & 55.96          & 62.26                         & 0.81                           \\
\textbf{ECO}     &  34.24    &  17.45  &  63.52  & 97.26 &  98.60 &  45.62  & 90.16 & 49.74  & 60.72  &  88.51 &  42.07 & 55.12 & 62.84  &  2.43   \\
\textbf{GA}      & 73.24          & 83.41          & 51.10          & 95.03          & 97.55          & 47.02          & 92.30          & 43.19          & 55.75          & 87.61          & 42.34          & 55.22          & 61.18                         & 0.78                           \\
\textbf{GA+GDR}  & 75.02          & 85.13          & 50.15          & 96.20          & 98.36          & \textbf{47.25} & 92.30          & 42.82          & 55.74          & 87.46          & 41.51          & 54.84          & 60.97                         & 0.76                           \\
\textbf{GA+KLR}  & 80.92          & 89.73          & 51.26          & 96.90          & 98.14          & 46.91          & 93.30          & 44.60          & 57.47          & 88.03          & 43.03          & 55.99          & 62.15                         & 0.73                          \\
\textbf{DPO}     & 72.25          & 92.15          & 55.17          & 81.01          & 94.40          & 44.03          & 88.97          & \colorbox{lightblue!50}{48.64}          & \colorbox{lightblue!50}{62.75}          & 86.61          & 45.61          & 57.46          & 62.39                         & 0.76                           \\
\textbf{DPO+GDR} & 36.59          & 87.10          & \colorbox{lightblue!50}{56.27}          & 90.23          & 97.01          & 43.02          & 90.63          & 46.30          & 59.93          & 85.75          & 43.56          & 53.08          & 61.05                         & 0.99                           \\
\textbf{DPO+KLR} & 87.91          & 95.71          & 54.18          & 88.38          & 96.72          & 44.68          & 93.63          & 48.37          & 62.57          & 86.61          & 45.14          & 56.91          & \colorbox{lightpurple!50}{63.10}                         & 0.69                           \\
\textbf{NPO}     & 71.04          & 84.43          & 51.41          & 95.68          & 97.64          & 46.95          & 91.30          & 43.74          & 56.43          & 87.89          & 42.80          & 55.85          & 61.57                         & 0.79                           \\
 \textbf{NPO+GDR} & 71.33          & 84.41          & 51.44          & 95.74          & 97.71          & 46.90          & 92.30          & 43.84          & 56.46          & 87.89          & 42.75          & 55.61          & 61.60                         & 0.79                           \\
\textbf{NPO+KLR} & 71.46          & 84.86          & 51.37          & 95.66          & 97.68          & 46.94          & 92.30          & 43.97          & 56.75          & 87.89          & 42.97          & 55.75          & 61.74                         & 0.79                           \\
\textbf{TV}      & 68.29          & \colorbox{lightblue!50}{78.67}          & 52.55          & 94.39          & 96.57          & 46.30          & 93.30          & 44.03          & 57.31          & 89.17          & 43.64          & 56.20          & 61.92                         & 0.84                           \\
\textbf{WHP}     & 96.77          & 80.70          & 51.26          & 98.06          & 98.00          & 46.91          & \textbf{94.30} & 42.53          & 54.95          & 88.32          & 41.79          & 55.01          & 61.03                         & 0.69                           \\
\textbf{ULD}     & 94.45          & 97.78          & 50.73          & 95.86          & 97.81          & 46.80          & 92.77          & 45.41          & 58.73          & 88.75          & 44.10          & \colorbox{lightblue!50}{58.06}          & 62.93                         & 0.65                           \\
\midrule
\textbf{\ours}  &  34.91 & \colorbox{lightpurple!50}{\textbf{39.02}} & 62.63 & 82.78 & 84.67 & 42.21 & 91.50 & 50.27 & 65.18 & 88.75 & \colorbox{lightpurple!50}{\textbf{50.22}} & 59.60 & 63.51 & 1.72    \\
\textbf{\oursfast}    & \textbf{33.66} & 39.23 & \colorbox{lightpurple!50}{\textbf{64.39}} & 83.63          & 88.24          & 41.89          & 91.30           & \colorbox{lightpurple!50}{\textbf{52.57}} & \colorbox{lightpurple!50}{\textbf{68.21}} & \textbf{89.60}  & 49.77 & \colorbox{lightpurple!50}{\textbf{63.94}} & \colorbox{lightpurple!50}{\textbf{64.89}}                & \textbf{1.78}                  \\
\bottomrule
\end{tabular}
\end{small}
\end{center}
\vskip -0.1in
\end{table*}

\subsection{Perturbed Distribution Matching}
Building on \metric, we identify the top-$K$\% most sensitive tokens and inject random noise into their embeddings before feeding them into the model, resulting in the perturbed unlearning sample $x'$. We first pass the original unlearning sample $x$ into the unlearning model $f_{\theta_{u}}$ to obtain the clean-run next-token probability distribution $p(y|x)$, where $y = f_{\theta_{u}}(y|x)$ represents the model's output. Similarly, the corrupted-run next-token probability distribution $p(y|x')$ is obtained by feeding the perturbed sample $x'$ into $f_{\theta_{u}}$. As illustrated in Fig.~\ref{method}, the clean-run probability distribution $p(y|x)$ assigns the highest probabilities to tokens that are informative and \textit{fact-related}. In contrast, for the corrupted-run distribution $p(y|x')$, the model fails to recall related facts, resulting in token generation that relies primarily on context or grammar, with top-ranked tokens being \textit{fact-unrelated}.

Consequently, the corrupted-run probability distribution intuitively simulates a natural unlearning effect, allowing the model to behave as if it had never been trained on the given example. To replicate such a natural unlearning environment, we subtract $p(y|x)$ from $p(y|x')$, using a tuning coefficient $C$ to control the strength of forgetting: 
$p(Y_t|y_{<t}) = p(y|x') - C \cdot p(y|x).$
This subtraction suppresses the probabilities of \textit{fact-related} tokens, while maintaining high rankings for \textit{fact-unrelated} tokens. As a result, the distribution of irrelevant tokens is preserved, thereby maintaining the model’s utility.

We then achieve unlearning by fine-tuning $f_{\theta_{u}}$ to match the subtracted logit probability distribution $p(Y_t|y_{<t})$. For autoregressive text generation, this is decomposed into a step-wise KL divergence~\cite{f-Divergence}:
\begin{align}
L & = 
%E_{Y \sim p}\left[log \frac{p(\mathbf{Y})}{q_\theta(\mathbf{Y})}\right] \\ & = 
- \sum_{i=1}^{t} \sum_{Y_i \in V} p(Y_i|y_{<i}) log q_\theta (Y_i|y_{<i}),
\end{align}
where $V$ is the vocabulary and $q_\theta$ represents the predicted distributions of the unlearn model $f_{\theta_{u}}$. 
We provide a more comprehensive explanation of the algorithmic procedure presented in Alg.~\ref{alg:unlearning}.

\begin{algorithm}[H]
\caption{\ours: Perturbation-based Unlearning}
\label{alg:unlearning}
\begin{algorithmic}[1]
\renewcommand{\algorithmicrequire}{\textbf{Input:}}
\renewcommand{\algorithmicensure}{\textbf{Output:}}
\REQUIRE Unlearning sample $x$ with $m$ tokens, target model $f_{\theta_{u}}$, tuning coefficient $C$
\ENSURE Fine-tuned unlearning model $f_{\theta_{u}}$

\STATE Identify the top-$K$\% most sensitive tokens in $x$ using \metric.
\STATE Introduce random noise to the $K$\% tokens embeddings to obtain the perturbed sample: $x'$.
\STATE Compute the clean-run next-token probability distribution:
$p(y|x) = f_{\theta_{u}}(y|x).$
\STATE Compute the corrupted-run next-token probability distribution:
$
p(y|x') = f_{\theta_{u}}(y|x').
$
\STATE Subtract the clean-run distribution from the corrupted-run distribution to emulate forgetting:\\
$
p(Y_t|y_{<t}) = p(y|x') - C \cdot p(y|x).
$
\STATE Fine-tune $f_{\theta_{u}}$ to match $p(Y_t|y_{<t})$ by minimising the step-wise KL divergence:\\
$
L = - \sum_{i=1}^{t} \sum_{Y_i \in V} p(Y_i|y_{<i}) \log q_\theta(Y_i|y_{<i}),
$
where $V$ is the vocabulary and $q_\theta$ represents the predicted distributions of $f_{\theta_{u}}$.
\STATE Update $f_{\theta_{u}}$ using gradient descent to minimise $L$.
\renewcommand{\algorithmicensure}{\textbf{Return:}}
\ENSURE Fine-tuned unlearning model $f_{\theta_{u}}$.
\end{algorithmic}
\end{algorithm}

In general, the advantage of \ours~lies in its generality and simplicity. 
Regarding \emph{generality}, 
on the one hand, the corrupted token embeddings in the first layer prevent the model from recalling any facts related to the unlearning sample across subsequent middle layers. On the other hand, subtracting the clean-run probability distribution causes both the answer and answer-related tokens to drop significantly in the probability distribution. Consequently, the unlearned model fails to generate rephrased answers during inference. Moreover, \ours~preserves the distribution of irrelevant tokens, minimizing side effects and maintaining the model's utility~\cite{Hurt}. 
In terms of \emph{simplicity}, the identification of sensitive tokens is automated and \ours~requires no additional training of a reinforced model~\cite{RKLD,ULD} or scope classifier~\cite{eco}, making the training process more efficient.

\begin{table*}[]
% \vspace{-1mm}
\caption{Experimental results on the Rephrased TOFU dataset using \ttsmall{LLaMA2-7B}. Notably, the baseline methods struggle to generalise to rephrased unlearning samples. In contrast, \ours~outperforms the baselines by up to 40.73\% (53.37 $\rightarrow$ 31.63) in Probability and 15.67\% (60.75 $\rightarrow$ 70.27) in Truth Ratio.}
\vspace{-2mm}
\label{TOFU_rephrase}
\setlength{\tabcolsep}{3pt}
\vskip 0.15in
\begin{center}
\begin{small}
\begin{tabular}{c|ccccc|ccccc|ccccc}
\toprule
\textbf{Dataset} & \multicolumn{5}{c|}{\textbf{Rephrased Forget01 Dataset}} & \multicolumn{5}{c|}{\textbf{Rephrased Forget05 Dataset}} & \multicolumn{5}{c}{\textbf{Rephrased Forget10 Dataset}}  \\ 
\textbf{Metric} & \textbf{RG↓}   & \textbf{Pr↓}   & \textbf{TR↑}   & \textbf{MU↑}   & \textbf{FRT↑} & \textbf{RG↓}   & \textbf{Pr↓}   & \textbf{TR↑}   & \textbf{MU↑}   & \textbf{FRT↑} & \textbf{RG↓}   & \textbf{Pr↓}   & \textbf{TR↑}   & \textbf{MU↑}   & \textbf{FRT↑} \\
\midrule
\textbf{Retain}                   & 37.18          & 15.48          & 65.64          & 61.11          & 2.32          & 35.78          & 12.29          & 63.87          & 59.30          & 2.47          & 34.48          & 12.10          & 64.21          & 58.87          & 2.53          \\ 
\midrule
\textbf{ICL}                      & 44.34          & 68.98          & 55.19          & 62.26          & 1.10          & 44.13          & 64.23          & 52.29          & 62.26          & 1.15          & 43.48          & 64.53          & 52.84          & 62.26          & 1.15          \\
\textbf{ECO}  & 42.61  & 24.79  & 68.14  & 62.38 & 1.85          & 40.72  & 36.87 & 62.87 & 62.84 & 1.62  & 39.43 & 42.72 & 60.72 & 62.60  & 1.52 \\
\textbf{GA}                       & 43.38          & \colorbox{lightblue!50}{29.07}          & 57.55          & 60.41          & 1.67          & 46.04          & 60.21          & 50.65          & 61.18          & 1.15          & 44.32          & 59.26          & 52.40          & 60.87          & 1.18          \\
\textbf{GA+GDR}                   & 46.59          & 62.50          & 51.96          & 61.42          & 1.13          & 46.14          & 59.56          & 50.00          & 60.97          & 1.15          & 44.82          & 57.11          & 50.79          & 60.70          & 1.19          \\
\textbf{GA+KLR}                   & 46.37          & 62.06          & 52.85          & 61.94          & 1.14          & 48.21          & 64.08          & 50.73          & 62.15          & 1.11          & 47.81          & 64.14          & 51.24          & 61.96          & 1.11          \\
\textbf{DPO}                      & 27.48          & 60.45          & \colorbox{lightblue!50}{60.75}          & 63.42          & 1.44          & 41.56          & 69.28          & 55.13          & 62.39          & 1.13          & \textbf{30.50} & 68.51          & \colorbox{lightblue!50}{55.77}          & 60.32          & 1.22          \\
\textbf{DPO+GDR}                  & 29.39          & 64.84          & 59.46          & 63.03          & 1.34          & \textbf{27.76} & 65.40          & \colorbox{lightblue!50}{56.50}          & 61.05          & 1.31          & 36.73          & 66.37          & 53.90          & 60.23          & 1.17          \\
\textbf{DPO+KLR}                  & 30.99          & 66.45          & 59.60          & 63.58          & 1.31          & 45.61          & 70.61          & 54.02          & \textbf{63.10} & 1.09          & 40.57          & 70.70          & 55.42          & 61.45          & 1.10          \\
\textbf{NPO}                      & 44.08          & 30.88          & 57.51          & 60.57          & 1.62          & 47.34          & 61.48          & 50.88          & 61.57          & 1.13          & 44.31          & 60.68          & 52.98          & 61.52          & 1.17          \\
\textbf{NPO+GDR}                  & 43.74          & 30.77          & 57.48          & 60.47          & 1.62          & 46.87          & 61.53          & 50.86          & 61.60          & 1.14          & 44.46          & 60.78          & 52.89          & 61.63          & 1.17          \\
\textbf{NPO+KLR}                  & 44.84          & 30.95          & 57.63          & 60.49          & 1.60          & 47.17          & 61.84          & 50.86          & 61.74          & 1.13          & 44.58          & 60.73          & 52.95          & 61.65          & 1.17          \\
\textbf{TV}                       & 42.22          & 45.17          & 56.04          & 61.68          & 1.41          & 45.43          & 59.76          & 52.18          & 61.92          & 1.18          & 39.39          & 50.03          & 53.62          & 60.18          & 1.35          \\
\textbf{WHP}                      & 49.98          & 53.61          & 52.15          & 61.83          & 1.19          & 50.59          & \colorbox{lightblue!50}{53.37}          & 50.62          & 61.03          & 1.17          & 48.21          & 57.19          & 51.20          & 60.96          & 1.16          \\
\textbf{ULD}                      & 29.76          & 45.89          & 59.90          & 58.95          & 1.56          & 49.66          & 63.98          & 50.82          & 62.93          & 1.11          & 30.63          & 41.18          & 52.29          & 63.03          & 1.76          \\
\midrule
\textbf{\ours} & 27.19 & 17.91 & 70.27 & 64.30 & 2.85 & 29.63 & 31.63 & 63.09 & \textbf{63.51} & \textbf{2.07} & 33.56 & 40.91 & 62.70 & 64.40 & 1.73 \\
\textbf{\oursfast} & \textbf{26.69}          & \colorbox{lightpurple!50}{\textbf{14.75}}          & \colorbox{lightpurple!50}{\textbf{71.72}} & \textbf{65.06} & \textbf{3.14}          &28.96        & \colorbox{lightpurple!50}{\textbf{30.14}}         & \colorbox{lightpurple!50}{\textbf{68.26}}          & 61.21          & \textbf{2.07}          & 32.72          & \textbf{37.99} & \colorbox{lightpurple!50}{\textbf{65.96}}          & \textbf{64.80} & \textbf{1.83} \\
\bottomrule
\end{tabular}
\end{small}
\end{center}
\vskip -0.1in
\end{table*}

\begin{table*}[t]
\caption{Experimental results on the Rephrased Harry Potter dataset. \ours~demonstrates effective generalisation to rephrased forget data, achieving relative improvements of up to 20.36\% (84.39 $\rightarrow$ 67.20) in Rephrased Forget ROUGE and 22.72\% (81.73 $\rightarrow$ 63.16) in Rephrased Forget F1 when tested on \ttsmall{LLaMA2-7B}. }
\vspace{-2.5mm}
\label{Harry_rephrase}
\setlength{\tabcolsep}{2.3pt}
\begin{center}
\begin{small}
\begin{tabular}{c|ccc|ccc|cc||ccc|ccc|cc}
\toprule
\textbf{Model}                    & \multicolumn{8}{c||}{\textbf{Phi-1.3B}} & \multicolumn{8}{c}{\textbf{LLaMA2-7B}}   \\
\midrule
\textbf{Dataset}              & \multicolumn{3}{c|}{\textbf{Forget}}              & \multicolumn{3}{c|}{\textbf{Rephrased Forget}}    & \multirow{2}{*}{\textbf{MU↑}} & \multirow{2}{*}{\textbf{FRT↑}} & \multicolumn{3}{c|}{\textbf{Forget}}              & \multicolumn{3}{c|}{\textbf{Rephrased Forget}}    & \multirow{2}{*}{\textbf{MU↑}} & \multirow{2}{*}{\textbf{FRT↑}} \\
\textbf{Metric}               & \textbf{RG↓}   & \textbf{Pr↓}   & \textbf{F1↓}   & \textbf{RG↓}   & \textbf{Pr↓}   & \textbf{F1↓}   &                               &                                & \textbf{RG↓}   & \textbf{Pr↓}   & \textbf{F1↓}   & \textbf{RG↓}   & \textbf{Pr↓}   & \textbf{F1↓}   &                               &                                \\
\midrule
\textbf{Retain}               & 44.61          & 14.34          & 44.49          & 43.55          & 14.10          & 43.74          & 62.73                         & 1.84                           & 43.84          & 19.43          & 39.89          & 41.11          & 19.58          & 36.40          & 83.99                         & 2.52                           \\
\midrule
\textbf{ICL}                  & 71.15          & 45.18          & 68.35          & 62.67          & 36.06          & 61.96          & 61.25                         & 1.06                           & 100.00         & 99.68          & 100.00         & 98.17          & 94.81          & 96.93          & \textbf{88.19}                & 0.90                           \\
\textbf{GA}                   & 77.08          & 49.93          & 73.68          & 66.92          & 40.35          & 64.73          & 64.70                         & 1.04                           & 93.53          & 68.72          & 91.73          & 93.10          & 66.10          & 90.81          & 82.45                         & 0.98                           \\
\textbf{GA+GDR}               & 73.64          & 45.27          & 69.79          & 64.15          & 37.26          & 61.42          & \textbf{65.04}                & 1.11                           & 95.20          & 74.69          & 93.21          & 92.30          & 71.96          & 90.29          & 86.22                         & 1.00                           \\
\textbf{GA+KLR}               & 69.02          & 38.19          & 64.89          & 59.76          & 32.27          & 57.39          & 63.73                         & 1.19                           & 93.53          & 71.75          & 91.73          & 93.10          & 68.54          & 90.81          & 83.13                         & 0.98                           \\
\textbf{DPO}                  & 75.86          & 48.66          & 73.62          & 68.75          & 38.74          & 67.82          & 62.76                         & 1.01                           & 92.21          & 76.67          & 89.85          & 89.77          & 73.54          & 86.71          & 83.02                         & 0.98                           \\
\textbf{DPO+GDR}              & 80.23          & 58.19          & 77.46          & 73.19          & 45.15          & 71.03          & 64.82                         & 0.96                           & 91.21          & 81.61          & 88.38          & 88.87          & 77.98          & 84.55          & 82.27                         & 0.96                           \\
\textbf{DPO+KLR}              & 75.93          & 49.68          & 73.95          & 70.09          & 39.44          & 69.15          & 63.21                         & 1.00                           & 91.38          & 68.50          & 89.09          & 87.93          & 65.78          & 83.65          & 81.34                         & 1.00                           \\
\textbf{NPO}                  & 68.69          & 38.42          & 64.56          & 59.93          & 32.44          & 57.39          & 63.97                         & 1.19                           & 92.53          & \colorbox{lightblue!50}{65.05}          & 90.01          & 91.70          & \colorbox{lightblue!50}{63.06}          & 88.99          & 81.72                         & 1.00                           \\
\textbf{NPO+GDR}              & 69.11          & 38.62          & 64.94          & 60.55          & 32.53          & 58.18          & 64.32                         & 1.19                           & 93.53          & 67.56          & 91.01          & 91.70          & 65.44          & 88.99          & 82.63                         & 1.00                           \\
\textbf{NPO+KLR}              & 68.69          & 38.39          & 64.56          & 60.21          & 32.35          & 57.70          & 63.89                         & 1.19                           & 92.53          & 65.18          & 90.01          & 91.70          & 63.19          & 88.99          & 81.76                         & 1.00                           \\
\textbf{TV}                   & 79.58          & 55.78          & 76.27          & 70.82          & 44.30          & 68.06          & 64.87                         & 0.99                           & 92.47          & 69.56          & 89.94          & 91.97          & 66.19          & 89.62          & 82.42                         & 0.99                           \\
\textbf{WHP}                  & 71.72          & 40.94          & 66.87          & 62.31          & 31.38          & 60.06          & 64.52                         & 1.16                           & \colorbox{lightblue!50}{87.35}          & 73.20          & \colorbox{lightblue!50}{84.90}          & \colorbox{lightblue!50}{84.39}          & 70.59          & \colorbox{lightblue!50}{81.73}          & 86.88                         & 1.08                           \\
\textbf{ULD}                  & 88.35          & 71.79          & 85.84          & 75.04          & 50.50          & 72.84          & 61.18                         & 0.83                           & 89.75          & 73.24          & 85.31          & 88.83          & 69.84          & 82.65          & 81.65                         & 1.00                           \\
\midrule
\textbf{\ours} & \colorbox{lightpurple!50}{\textbf{56.52}} & \textbf{31.41} & \colorbox{lightpurple!50}{\textbf{59.53}} & \textbf{54.25} & \textbf{29.22} & \textbf{54.98} & 62.10 & \textbf{1.30} & \colorbox{lightpurple!50}{\textbf{70.15}} & \colorbox{lightpurple!50}{\textbf{55.17}} & \colorbox{lightpurple!50}{\textbf{68.85}} & \colorbox{lightpurple!50}{\textbf{67.20}} & \colorbox{lightpurple!50}{\textbf{54.30}} & \colorbox{lightpurple!50}{\textbf{63.16}} & 82.49 & \textbf{1.31}  \\
\textbf{\oursfast} & 65.21 & 35.75 & 63.02 & 59.11 & 31.24 & 56.02 & 63.20                         & 1.22                  & 71.88 & 56.36 & 69.54 & 69.89 & 56.59 & 65.49 & 82.65   & 1.27                               
      \\
\bottomrule
\end{tabular}
\end{small}
\end{center}
% \vskip -0.2in
\end{table*}

\section{Evaluation of \ours}
In this section, we evaluate the effectiveness of \ours~and analyze the results. The implementation parameters of \ours~are consistent with those of the baselines. We integrate GDR with $RW=1$, set the percentage of perturbed tokens to $K = 0.4$, and maintain perturbation ratio $P = 0.4$, coefficient $C=0.1$.

\begin{figure}[t]
\centering
\includegraphics[width=0.48\textwidth]{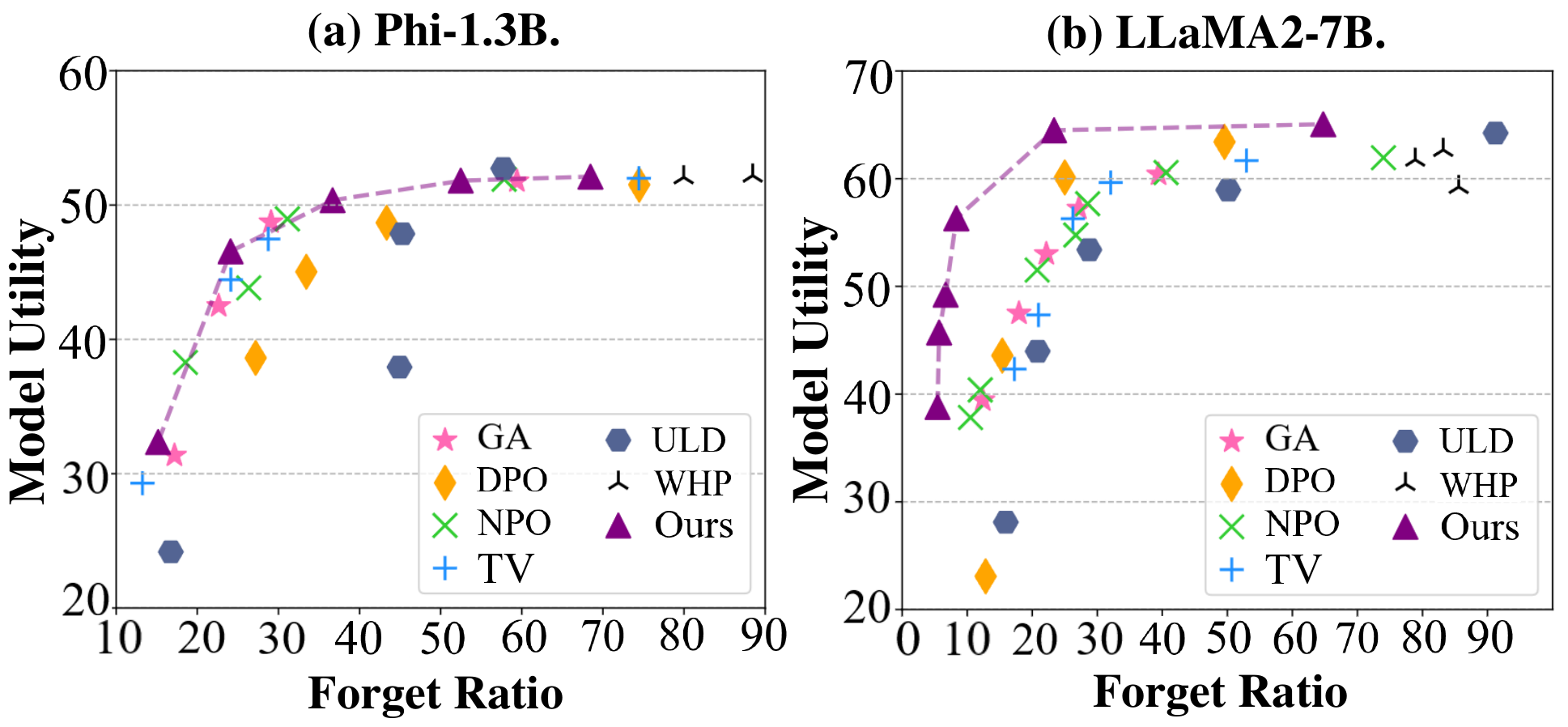}
\caption{The curves illustrating how Model Utility changes with the Forget Ratio. The closer a method is to the upper left corner, the better it balances model utility and the forgetting effect. The proposed \ours~encompasses nearly all baseline methods from the top left, demonstrating superior unlearning performance.
}
\vspace{-0.15in}
\label{curve}
\end{figure}

\subsection{Improved Unlearning Capabilities} 
When tested on TOFU Forget05 dataset using \ttsmall{LLaMA2-7B}, as shown in Tab.~\ref{tofu_llama_full}, \ours~demonstrates remarkable effectiveness in unlearning knowledge. Specifically, \ours~outperforms the baselines with a relative improvement of up to \textbf{50.40\%} (78.67 $\rightarrow$39.02) in Forget Probability and 11.3\% (56.27$\rightarrow$62.63) on Forget Truth Ratio. 

When tested on the Harry Potter dataset, as shown in Tab.~\ref{Harry_rephrase}, \ours~consistently outperforms baselines on both \ttsmall{Phi-1.3B} and \ttsmall{LLaMA2-7B}. Specifically, \ours~achieves an absolute improvement of up to 17.2\% (87.35 $\rightarrow$ 70.15) in Forget ROUGE, 9.88\% (65.05 $\rightarrow$ 55.17) in Forget Probability and 16.05\% (84.90 $\rightarrow$ 68.85) in Forget F1 when evaluated on \ttsmall{LLaMA2-7B}, all while maintaining high model utility. Furthermore, the performance improvements of \ours~are more pronounced on \ttsmall{LLaMA2-7B} compared to \ttsmall{Phi-1.3B}, highlighting its resilience to model scaling and its potential for application to larger models.

This significant reduction in Forget Probability can be attributed to \ours's ability to achieve unlearning at the token probability distribution level, effectively lowering the probabilities of the correct answer and its related tokens. Specifically, we plot the curve in Fig.~\ref{curve} based on the TOFU dataset, illustrating how model utility changes with the Forget Ratio, calculated as the mean of Forget ROUGE and Forget Probability. The closer a method is to the upper-left corner, the better it balances model utility and the unlearning effect. \ours~encompasses nearly all baseline methods from the top left, demonstrating superior unlearning performance.

\begin{table*}[t]
\caption{Experimental results for inverted relation data, subject-replaced data, and one-hop reasoned data on the ZsRE dataset using \ttsmall{Phi-1.3B}. Most methods struggle to forget the “hard” in-scope knowledge that are logically related to the original unlearning sample. Nevertheless, \ours~achieves an improvement of up to 9.41\% on FRT ratio (0.85$\rightarrow$0.93). }
\label{zsre_rephrase}
\vspace{-1mm}
\setlength{\tabcolsep}{3.2pt}
\vskip 0.13in
\begin{center}
\begin{small}
\begin{tabular}{c|ccccc|ccccc|ccccc}
\toprule
\multicolumn{1}{c|}{\textbf{Dataset} }                 & \multicolumn{5}{c|}{\textbf{Inversed Relation}}                                     & \multicolumn{5}{c|}{\textbf{Subject Replacement}}                                   & \multicolumn{5}{c}{\textbf{One-Hop Reasoning}}                                     \\
\multicolumn{1}{c|}{\textbf{Metric}}                   & \textbf{RG↓}   & \textbf{Pr↓}   & \textbf{F1↓}   & \textbf{MU↑}   & \textbf{FRT↑}  & \textbf{RG↓}   & \textbf{Pr↓}   & \textbf{F1↓}   & \textbf{MU↑}   & \textbf{FRT↑}  & \textbf{RG↓}   & \textbf{Pr↓}   & \textbf{F1↓}   & \textbf{MU↑}   & \textbf{FRT↑}  \\
\midrule
\textbf{Retain}                   & 50.43          & 17.04          & 50.08          & 70.04          & 1.79          & 50.43          & 17.04          & 50.08          & 70.04          & 1.79          & 49.09          & 19.10          & 49.01          & 70.89          & 1.81          \\
\midrule
\textbf{ICL}                      & 87.23          & 77.92          & 84.86          & 59.30          & 0.71          & \textbf{64.84} & \textbf{48.81} & \textbf{63.93} & 57.06          & \colorbox{lightpurple!50}{\textbf{0.96}} & \textbf{81.70} & 68.27          & \textbf{80.46} & 59.80          & \colorbox{lightblue!50}{0.78}          \\
\textbf{GA}                       & 84.46          & 69.92          & 83.47          & 64.36          & 0.81          & 92.35          & 83.25          & 92.40          & 68.10          & 0.76          & 98.76          & 91.09          & 98.28          & 66.11          & 0.69          \\
\textbf{GA+GDR}                   & 89.39          & 76.81          & 88.98          & 67.32          & 0.79          & 93.26          & 86.22          & 93.31          & 69.11          & 0.76          & 91.87          & 80.17          & 91.06          & 62.01          & 0.71          \\
\textbf{GA+KLR}                   & 85.33          & 69.88          & 84.61          & 64.30          & 0.80          & 91.67          & 83.11          & 91.72          & 68.11          & 0.77          & 98.85          & 91.14          & 98.38          & 66.01          & 0.69          \\
\textbf{DPO}                      & 82.54          & 63.80          & \colorbox{lightblue!50}{81.28}          & 62.13          & 0.82          & 96.46          & 84.94          & 96.51          & 68.75          & 0.74          & 98.58          & 92.52          & 98.14          & 68.54          & 0.71          \\
\textbf{DPO+GDR}                  & \colorbox{lightblue!50}{81.76}          & 66.19          & 81.46          & 63.09          & 0.83          & 97.15          & 87.25          & 97.20          & \textbf{69.35} & 0.74          & 97.62          & 91.13          & 97.14          & 68.82          & 0.72          \\
\textbf{DPO+KLR}                  & 88.31          & 69.66          & 87.54          & 64.68          & 0.79          & 96.46          & 85.69          & 96.51          & 68.85          & 0.74          & 98.77          & 92.67          & 98.34          & 68.28          & 0.71          \\
\textbf{NPO}                      & 85.28          & 68.92          & 84.34          & 64.06          & 0.81          & 92.12          & 82.67          & 92.17          & 68.02          & 0.76          & 99.25          & 92.88          & 98.83          & 67.35          & 0.69          \\
\textbf{NPO+GDR}                  & 86.66          & 70.32          & 85.50          & 64.79          & 0.80          & 92.35          & 83.83          & 92.40          & 68.58          & 0.77          & 99.30          & 93.61          & 98.89          & 67.84          & 0.70          \\
\textbf{NPO+KLR}                  & 85.80          & 68.89          & 84.86          & 64.09          & 0.80          & 92.12          & 82.88          & 92.17          & 68.01          & 0.76          & 99.25          & 92.94          & 98.83          & 67.37          & 0.69          \\
\textbf{TV}                       & 96.19          & 83.95          & 95.66          & 68.57          & 0.75          & 85.87          & 79.56          & 84.93          & 67.29          & \colorbox{lightblue!50}{0.81}          & 98.50          & 92.82          & 98.10          & \textbf{69.22} & 0.72          \\
\textbf{WHP}                      & 96.76          & 86.34          & 96.23          & \textbf{68.62} & 0.74          & 91.44          & 83.58          & 91.55          & 68.91          & 0.78          & 99.36          & 94.82          & 98.95          & 69.15          & 0.71          \\
\textbf{ULD}                      & 85.62          & \textbf{54.75} & 84.89          & 64.11          & \colorbox{lightblue!50}{0.85}          & 96.69          & 81.04          & 96.74          & 64.84          & 0.71          & 91.23          & \textbf{67.71} & 90.10          & 62.78          & 0.76          \\
\midrule
\textbf{\ours} & \colorbox{lightpurple!50}{\textbf{73.95}} & 55.34 & \colorbox{lightpurple!50}{\textbf{72.52}} & 62.54 & \colorbox{lightpurple!50}{\textbf{0.93}} & 86.53 & 75.29 & 85.51 & 67.13 & \colorbox{lightblue!50}{0.81} & 85.63 & 68.05 & 83.68 & 63.83 & \colorbox{lightpurple!50}{\textbf{0.81}} \\ 
\textbf{\oursfast} & 77.00 & 56.51   & 74.87 & 63.76          & 0.92 & 88.19          & 79.62          & 87.16          & 68.56          & \colorbox{lightblue!50}{0.81}          & 85.95          & 70.34          & 83.74          & 64.77          & \colorbox{lightpurple!50}{\textbf{0.81}} \\
\bottomrule
\end{tabular}
\end{small}
\end{center}
% \vskip -0.15in
\end{table*}

\begin{table*}[!htb]
\centering
\footnotesize
\caption{Although existing methods can effectively forget the original unlearning samples, they fail to generalise to forget rephrased unlearning samples (text in red). In contrast, \ours~demonstrates superior generalisation ability (text in green).}
% \vskip 0.05in
\resizebox{\textwidth}{!}{
\begin{tabular}{l}
\toprule
\toprule
{\textbf{The Underlying Problem: Failure to Generalise to Rephrased Unlearning Samples}} \\  
\toprule
\toprule
\begin{tabular}[c]{@{}l@{}}
{\textbf{Dataset}: TOFU} \\ 
{\textbf{Unlearning Sample:}}
What genre is author Basil Mahfouz Al-Kuwaiti most known for in his writing? \\ 
{\textbf{Rephrased Unlearning Sample:}}
For which genre of literature is Basil Mahfouz Al-Kuwaiti best recognized? \\
\textbf{Ground Truth:} 
Basil Mahfouz Al-Kuwaiti is most known for his writings in the \colorbox{gray!50}{French literature genre}. \\
\toprule
\textbf{Prediction of ULD on the Unlearning Sample: }\\
The genre that author Basil Mahfouz Al-Kuwaiti is best known for is {\color[rgb]{0,0.5,0} the travelogue genre}. \\
\textbf{Prediction of ULD on the Rephrased Unlearning Sample: }\\
Basil Mahfouz Al-Kuwaiti is best known for his contributions to {\color{red} the French literature genre}. \\
\midrule
\textbf{Prediction of \ours~on the Unlearning Sample: }\\
Basil Mahfouz Al-Kuwaiti is primarily known for writing {\color[rgb]{0,0.5,0} in the genre of erotica}. \\
\textbf{Prediction of \ours~on the Rephrased Unlearning Sample: }\\
Basil Mahfouz Al-Kuwaiti is best known for writing books in {\color[rgb]{0,0.5,0} the genre of mythology}.\\
\bottomrule
\bottomrule
\end{tabular}
\end{tabular}
}
\label{rephrase_question}
\end{table*}

\subsection{Improved Generalisation Capabilities}

When tested on the Rephrased TOFU dataset using \ttsmall{LLaMA2-7B} (Tab.~\ref{TOFU_rephrase}),  \ours~outperforms the baselines relatively by up to \textbf{40.73\%} (53.37 $\rightarrow$ 31.63) in Forget Probability, and 15.67\% (60.75 $\rightarrow$ 70.27) in Truth Ratio. At the same time, \ours~preserves knowledge on the Real World dataset better than the baselines when using \ttsmall{LLaMA2-7B}, even surpassing the retain model by up to 10.7\% (39.52 $\rightarrow$ 50.22) on Real World Probability and 6.84\% (52.76 $\rightarrow$ 59.60) on Real World Truth Ratio (Tab.~\ref{tofu_llama_full}). We present a case study of the generalisation issue. As shown in Tab.~\ref{rephrase_question}, the predictions generated by ULD remain highly similar to the ground truth (highlighted in red), reflecting suboptimal generalization. In contrast, \ours~produces entirely different responses from the ground truth, demonstrating superior generalization ability.

Furthermore, \ours~generalises effectively to rephrased Harry Potter data, achieving relative improvements of up to 20.36\% (84.39 $\rightarrow$ 67.20) in ROUGE and 22.72\% (81.73 $\rightarrow$ 63.16) in F1 when tested on \ttsmall{LLaMA2-7B} (Tab.~\ref{Harry_rephrase}), while incurring only a minor 1.5\% (83.99 $\rightarrow$ 82.49) reduction in model utility compared to the retain model.

\begin{figure}[t]
\begin{center}
\includegraphics[width=0.48\textwidth]{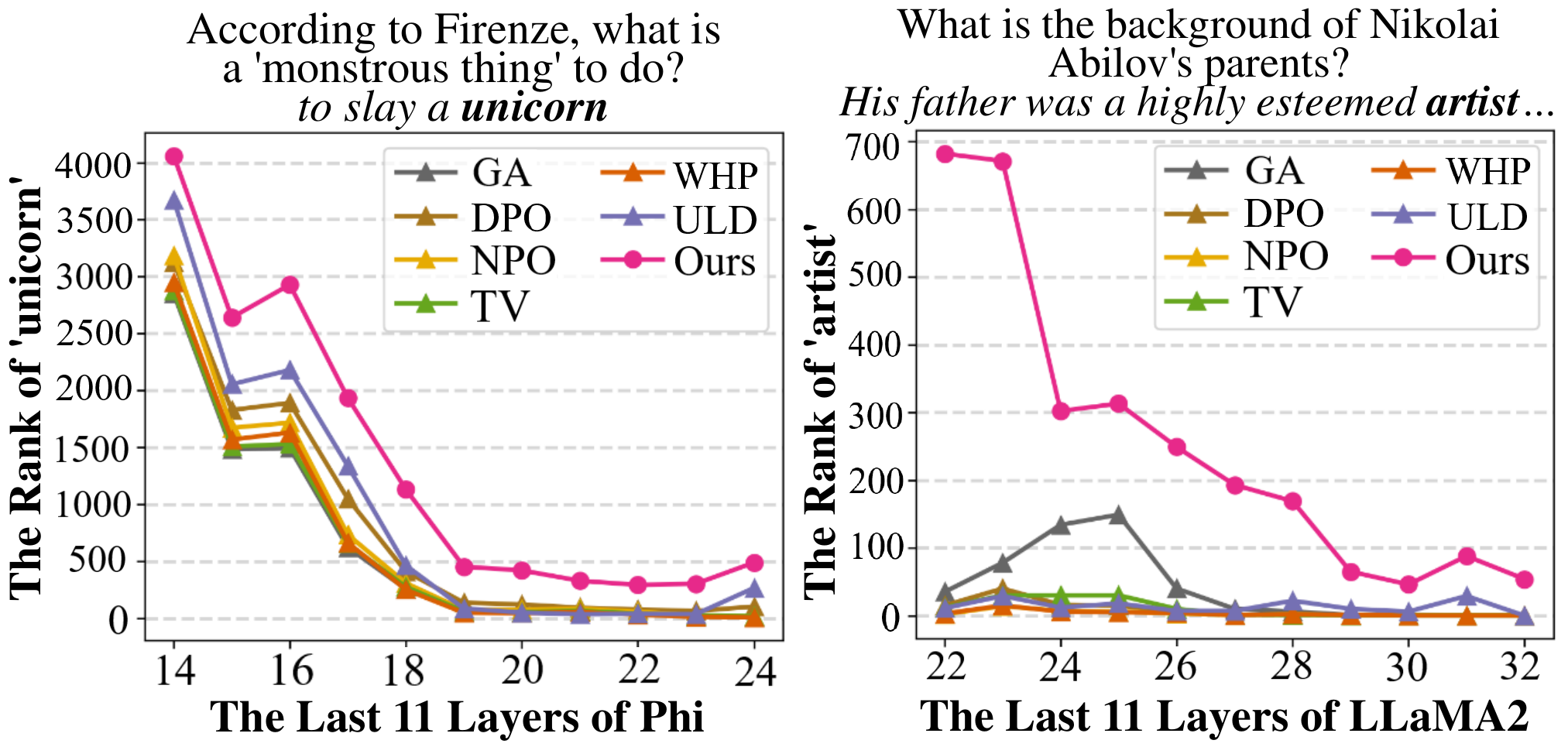}
\caption{The ranking of the first answer token in the next-token probability distribution across layers of the unlearned model. \ours~consistently achieves lower ranks across all layers, indicating that the model fails to recall facts.}
\label{layer append}
\end{center}
\vskip -0.15in
\end{figure}

When evaluated on the ZsRE dataset, as shown in Tab.~\ref{zsre_rephrase}, although the ICL-based method demonstrates superior unlearning performance, it suffers from lower model utility, which limits its applicability in real-world scenarios. Apart from it, \ours~achieves the best FRT ratio across all three subsets, particularly excelling on inverted relation data, where it shows a relative improvement of up to 9.41\% (0.85$\rightarrow$0.93) on FRT ratio. This highlights \ours's ability to strike a better balance between unlearning and retaining knowledge. However, the absolute generalisation performance of all methods in forgetting knowledge logically related to the original unlearning samples remains suboptimal, particularly when it comes to unlearning knowledge that requires one-hop reasoning, showing a gap of up to 2.23 times lower than the Retain model. Enhancing the absolute generalisation ability of the unlearned model is crucial in machine unlearning, leaving room for further exploration by the community.

\begin{figure*}[t]
\begin{center}
\includegraphics[width=0.98\textwidth]{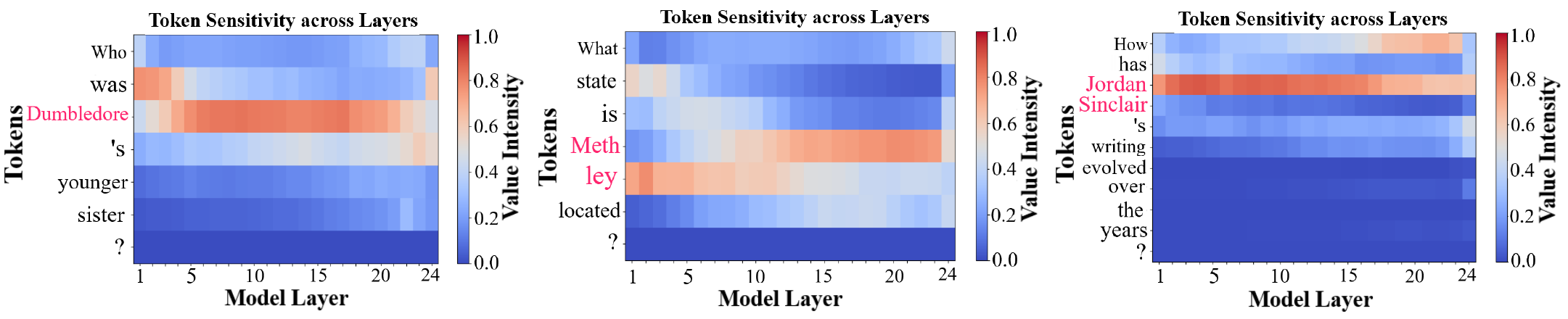}
\caption{Visualization of the model's sensitivity to each token across all layers. Subject words are highlighted with brighter colors and exhibit higher \metric~values, indicating the model's increased sensitivity to these tokens.}
\label{subjects_visual}
\end{center}
\vskip -0.1in
\end{figure*}

\subsection{Discussions}

We investigate whether \ours~effectively addresses the two underlying problems highlighted in Section~\ref{Benchmark Analysis}. For the first problem, as shown in Fig.~\ref{layer append}, the ranking of the first answer token in the next-token probability distribution of \ours~is significantly lower than that of other methods across layers, indicating that the unlearned model fails to recall the facts during inference, thereby \textbf{\textit{addressing Cause 1}}. As for the second problem, \ours~produces a lower probability of generating rephrased answers on both unlearning samples and rephrased unlearning samples (as shown in Tab.~\ref{cause2}), with relative reductions of up to 4.1\% and 9.4\%, respectively. Moreover, the probability delta is 33.3\% lower than that of other methods, effectively \textbf{\textit{addressing Cause 2}}. These results confirm that \ours~exhibits a much more generalised unlearning capability.

\begin{table}[h]
\caption{Comparison between the average \metric~values of subject words and the other words. Subject words exhibit higher \metric~values than other tokens, indicating greater sensitivity.}
\label{eigen_table}
\vspace{-0.15in}
\begin{center}
\begin{sc}
\begin{small}
\begin{tabular}{c|ccc}
\toprule
\textbf{Dataset}   & \textbf{TOFU} & \textbf{Harry} & \textbf{ZsRE} \\ 
\midrule
\textbf{Subject Words} & 0.000561      & 0.003219       & 0.005961      \\ 
% \midrule
\textbf{Other Words}   & 0.000219      & 0.001708       & 0.003082      \\ 
% \midrule
\textbf{Ratio}         & 2.66          & 1.86           & 1.93          \\ \bottomrule
\end{tabular}
\end{small}
\end{sc}
\end{center}
\vskip -0.2in
\end{table}

\subsection{Fast Alternative Implementation of \ours}
\label{Fast Implementation}
To reduce the computational overhead of identifying the top-$K$\% most sensitive tokens for each unlearning sample, we explore a more efficient implementation of \ours. We begin by computing the \metric~to determine the types of tokens to which the model is most sensitive. Experiments are conducted using \texttt{Phi-1.3B}.

As shown in Tab.~\ref{eigen_table}, the mean \metric~value of the subject words is up to 2.66 times higher than that of other words. We further visualize this by normalizing the \metric~values of all tokens in each sentence to a range between 0 and 1. As depicted in Fig.~\ref{subjects_visual}, the subject words consistently exhibit higher intensity values in the middle layers, indicating that \textbf{the model is more sensitive to subject tokens}. This observation aligns with the role of subject words in the mid-layer modules of generative language models, which are primarily responsible for recalling factual information~\cite{ROME,Recall_of_Factual}, thereby influencing the model's output to reflect memorized knowledge. It also corresponds to the structure of multi-hop reasoning, which typically begins with subject enrichment~\cite{Dissecting}. These insights confirm that subject tokens are key carriers of factual memory.

Based on this analysis, we propose a fast variant of our method, \oursfast, which directly perturbs all subject tokens in the unlearning sample while keeping all other hyper-parameters unchanged. This approach not only effectively suppresses the model’s ability to recall facts related to subjects—particularly efficient in copyright and privacy-related datasets—but also significantly reduces training time. Specifically, for the TOFU and ZsRE datasets, where subject words are already labeled, we directly locate these tokens in the unlearning samples and apply perturbation. For the Harry Potter dataset, which contains numerous character names, we treat character names as subject words and perturb them accordingly. In contrast, for WMDP and MUSE, where subject words are either not apparent or are scattered across long-form sentences, we revert to the original \ours~method for unlearning. 

We present the experimental results of \oursfast. As shown in Tab.\ref{TOFU_rephrase}, \oursfast~achieves unlearning performance comparable to existing methods, even surpasses \ours~by up to 17.64\% (17.91 $\rightarrow$ 14.75) on Rephrased TOFU Forget01 Probability, and 8.19 (63.09 $\rightarrow$ 68.26) on Rephrased TOFU Forget05 TR. Although \oursfast~underperforms \ours~on Harry Potter dataset, it still outperforms baseline methods, achieving relative improvements of up to 15.47\% (87.35 $\rightarrow$ 71.88) in Forget ROUGE, 8.69\% (65.05 $\rightarrow$ 56.36) in Forget Probability and 15.36\% (84.90 $\rightarrow$ 69.54) in Forget F1 when evaluated on \ttsmall{LLaMA2-7B}, all while maintaining high model utility. These results demonstrate the effectiveness of \oursfast~in achieving efficient forgetting with minimal computational overhead.

\begin{table*}[h]
\caption{The Fluency ratio on TOFU and ZSRE dataset using \ttsmall{LLaMA2-7B}. \ours~achieves the best Fluency ratio across all datasets, indicating better generation quality. The case study can be found in Tab.~\ref{Low-Quality-Generation}.}
\label{fluency}
\setlength{\tabcolsep}{2.3pt}
\begin{center}
\begin{small}
\begin{tabular}{c|cccccccccccccc}
\toprule
\textbf{Dataset}                  & \multicolumn{14}{c}{\textbf{TOFU - Forget05}} \\
\textbf{Metric}                   & \textbf{$\mathbf{\mathrm{ICL}}$}  & \textbf{$\mathbf{\mathrm{GA}}$} & \textbf{$\mathbf{\mathrm{GA}_{GD}}$} & \textbf{$\mathbf{\mathrm{GA}_{KL}}$} & \textbf{$\mathbf{\mathrm{DPO}}$} & \textbf{$\mathbf{\mathrm{DPO}_{GD}}$} & \textbf{$\mathbf{\mathrm{DPO}_{KL}}$} & \textbf{$\mathbf{\mathrm{NPO}}$} & $\mathbf{\mathrm{NPO}_{GD}}$ & \textbf{$\mathbf{\mathrm{NPO}_{KL}}$} & \textbf{$\mathbf{\mathrm{TV}}$}   & \textbf{$\mathbf{\mathrm{WHP}}$} & \textbf{$\mathbf{\mathrm{ULD}}$}  & \ours \\ \midrule
\textbf{Real Authors}    & 3.64          & 3.62        & 3.61            & 3.62            & 3.54         & 3.48             & 3.58             & 3.61         & 3.60             & 3.61             & \colorbox{lightblue!50}{3.70}          & 3.66         & 3.68          & \colorbox{lightpurple!50}{\textbf{4.29}} \\
\textbf{Real World}      & 3.87          & 3.88        & 3.83            & 3.86            & 3.62         & 3.69             & 3.67             & 3.90         & 3.87             & 3.88             & 3.97          & 3.89         & \colorbox{lightblue!50}{4.05}          & \colorbox{lightpurple!50}{\textbf{4.62}} \\
\textbf{Retain}          & 4.64          & 4.62        & 4.63            & 4.63            & 4.26         & 4.53             & 4.40             & 4.63         & 4.63             & 4.63             & 4.65          & 4.64         & 4.65          & \textbf{4.79} \\
\textbf{Forget}          & 4.58          & 4.67        & 4.68            & 4.68            & 4.26         & 3.37             & 4.56             & 4.67         & 4.67             & 4.66             & \textbf{4.72} & 4.72         & 4.72          & 4.72          \\
\textbf{Rephrased Forget} & 4.67          & 4.75        & 4.69            & 4.76            & 4.27         & 3.57             & 4.43             & 4.77         & 4.74             & 4.77             & 4.81          & 4.76         & 4.78          & \textbf{4.83} \\
\colorbox{gray!40}{\textbf{Average}}             & 4.28          & 4.31        & 4.29            & 4.31            & 3.99         & 3.73             & 4.13             & 4.32         & 4.30             & 4.31             & 4.37          & 4.33         & 4.37          & \textbf{4.65} \\ 
\midrule
\textbf{Dataset}                 & \multicolumn{14}{c}{\textbf{ZsRE - Inversed Relation}} \\ 
\midrule
\textbf{Real Authors}    & 0.10          & 0.10        & 0.10            & 0.10            & 0.10         & \textbf{0.23}    & 0.18             & 0.10         & 0.10             & 0.10             & 0.10          & 0.10         & 0.10          & 0.10          \\
\textbf{Real World}      & 0.16          & 0.16        & 0.15            & 0.15            & 0.24         & \textbf{0.32}    & 0.43             & 0.15         & 0.15             & 0.16             & 0.17          & 0.16         & 0.19          & 0.16          \\
\textbf{Retain}          & 0.44          & 0.44        & 0.44            & 0.44            & 0.44         & 0.44             & 0.44             & 0.44         & 0.44             & 0.44             & 0.44          & 0.44         & 0.44          & \textbf{0.49} \\
\textbf{Forget}          & 0.48          & 0.50        & 0.50            & 0.50            & 0.48         & 0.48             & 0.46             & \colorbox{lightblue!50}{0.51}      & \colorbox{lightblue!50}{0.51}      & \colorbox{lightblue!50}{0.51}             & 0.49          & 0.49         & 0.48          & \colorbox{lightpurple!50}{\textbf{1.00}} \\
\textbf{Rephrased Forget} & 0.48          & \colorbox{lightblue!50}{0.51}        & \colorbox{lightblue!50}{0.51}            & 0.49            & 0.48         & 0.49             & 0.48             & \colorbox{lightblue!50}{0.51}         & \colorbox{lightblue!50}{0.51}             & \colorbox{lightblue!50}{0.51}             & 0.49          & 0.49         & 0.48          & \colorbox{lightpurple!50}{\textbf{0.98}} \\
\colorbox{gray!40}{\textbf{Average}}             & 0.33          & 0.34        & 0.34            & 0.34            & 0.35         & 0.39             & 0.40             & 0.34         & 0.34             & 0.34             & 0.34          & 0.34         & 0.34          & \textbf{0.55} \\
\bottomrule
\end{tabular}
\end{small}
\end{center}
% \vskip -0.2in
\end{table*}

\section{Comprehensive Study}
In this section, we conduct a comprehensive study to evaluate the effectiveness of \ours. This includes assessing the quality of the unlearned model’s generations, evaluating performance on more diverse datasets and larger deep reasoning models, and analyzing computational overhead. Additionally, we perform ablation studies to investigate the impact of various hyper-parameters, including the percentage of perturbed tokens ($K$), the perturbation ratio ($P$), the tuning coefficient ($C$), discrete token-level perturbation, and the use of different retain losses.

\subsection{Evaluation on Generation Quality}
The impact of machine unlearning on language models is intricate, requiring a thorough and comprehensive evaluation to fully understand its effects. To this end, we perform additional tests to evaluate the generation quality of existing methods. Building on the previous work~\cite{ROME}, we introduce the Fluency metric to measure the fluency of the unlearned model’s output sentences. Fluency is measured by the weighted average of bi- and tri-gram entropies~\cite{entropies}, defined as $-\sum_k f(k)\log_2 f(k)$, where $f(\cdot)$ represents the $n$-gram frequency distribution. A higher Fluency score indicates more informative and diverse text generation. Experiments are conducted on \ttsmall{LLaMA2-7B} across all subsets. 

As shown in Tab.~\ref{fluency}, \ours~achieves a higher Fluency ratio on most datasets. Notably, when tested on the ZsRE dataset, \ours~outperforms others by nearly twofold on both the Forget dataset and the Rephrased Forget dataset. Moreover, \ours~achieves the highest average Fluency score across all datasets, with a relative improvement of up to 37.5\% (0.40$\rightarrow$0.55). We present a case study in Tab.~\ref{Low-Quality-Generation}, where Gradient Ascent generates low-quality outputs with repetitive content. In contrast, \ours~produces more fluent and readable sentences.

We hypothesize that this improvement arises because \ours~removes some noise—such as punctuation marks, delimiters, newlines, and other inconsequential tokens—from the next-token probability distribution through the probability subtraction process. As a result, the probability distribution becomes more refined, enhancing the model's generation quality. We present additional case studies of the unlearned model's outputs in the following section.

\begin{table*}[!htb]
\centering
\caption{Case study of the unlearned model's generation quality. The predictions from Gradient Ascent contain repetitive and redundant words (text in red). In contrast, \ours~produces more fluent and readable sentences (text in green).}
\resizebox{\textwidth}{!}{
\begin{tabular}{l}
\toprule
{\textbf{The Underlying Problem: Low-Quality Generation}} \\   
\toprule
\begin{tabular}[c]{@{}l@{}}
{\textbf{Dataset}: TOFU} \\ 
{\textbf{Unlearning Sample:}}
What was Hina Ameen's maiden book?   \\ 
\textbf{Ground Truth:} 
Hina Ameen's maiden book was \colorbox{gray!50}{``Manual of Mineralogy"}. \\
\toprule
\textbf{Prediction of Gradient Ascent: }\\
The maiden book of {\color{red} Hina Ameermunger Hina Ameer's maiden book} is {\color{red} `Ameermunger Hina Ameer's maiden book} is `Ameermunger'.  \\
\midrule
\textbf{Prediction of \ours: }\\
The maiden book of Hina was {\color[rgb]{0,0.5,0} `A Touch of Innocence'}. \\
\end{tabular}                                 \\
\bottomrule \\ 
\end{tabular}
}
\label{Low-Quality-Generation}
\end{table*}

\begin{table*}[]
\centering
\footnotesize
\caption{Experiment results on \ttsmall{DeepSeek-R1-Distill-Llama-8B} using the One-hop dataset. Deep reasoning models exhibit improved one-hop unlearning performance. However, the absolute gap between unlearning on the forget data and on the one-hop data remains substantial.}
% \vskip 0.1in
\label{deep reason}
\setlength{\tabcolsep}{8pt}
\begin{tabular}{c|ccc|ccc|ccc|c}
\toprule
\textbf{Dataset} & \multicolumn{3}{c|}{\textbf{Forget Data}}        & \multicolumn{3}{c|}{\textbf{Rephrased Forget Data}} & \multicolumn{3}{c|}{\textbf{One-hop Data}}       & \multirow{2}{*}{\textbf{MU}} \\
\textbf{Metric}  & \textbf{RG ↓}  & \textbf{Pr↓}   & \textbf{F1↓}   & \textbf{RG↓}    & \textbf{Pr↓}    & \textbf{F1↓}    & \textbf{RG ↓}  & \textbf{Pr↓}   & \textbf{F1↓}   &                              \\ \midrule
\textbf{GA+GD}   & 57.65          & 38.93          & 53.31          & 59.87           & 40.30           & 55.88           & 75.23          & 58.75          & 72.50          & 59.69                        \\
\textbf{DPO+GD}  & 59.15          & 42.89          & 58.20          & 60.61           & 42.70           & \textbf{53.04}  & \textbf{72.30} & 56.05          & 73.63          & 60.28                        \\
\textbf{NPO+GD}  & 57.52          & 35.19          & 52.87          & 58.91           & 39.89           & 53.07           & 74.23          & \textbf{54.95} & \textbf{72.32} & 59.85                        \\
\textbf{TV}      & 62.52          & 42.60          & 67.46          & 67.29           & 42.70           & 55.19           & 76.33          & 56.95          & 74.18          & \textbf{61.20}               \\
\textbf{ULD}     & 58.30          & 37.46          & 53.16          & 60.09           & 36.98           & 55.72           & 75.25          & 55.21          & 72.83          & 60.34                        \\
\textbf{WHP}     & 60.37          & 37.71          & 57.98          & 61.80           & 37.29           & 55.91           & 75.42          & 55.15          & 73.07          & 59.36                        \\
\textbf{RMU}     & 57.40          & 39.51          & 55.06          & \colorbox{lightblue!50}{57.66}           & 36.26           & 56.19           & 72.33          & 55.43          & 74.33          & 59.24                        \\
\textbf{\ours}   & \textbf{56.21} & \textbf{34.95} & \textbf{52.38} & \colorbox{lightpurple!50}{\textbf{54.95}}  & \textbf{34.58}  & 53.10           & 73.82          & 55.30          & 72.58          & 58.78                        \\ \bottomrule
\end{tabular}
\end{table*}

\begin{table*}[]
\centering
% \begin{}
\caption{Evaluation on the WMDP and MUSE datasets using \ttsmall{Vicuna-13B} and \ttsmall{MUSE-7B}, respectively. 
\ours~consistently achieves the highest FRT score, with improvements of up to 10.77\% (from 0.65 to 0.72).
}
% \vskip 0.05in
% \renewcommand{\arraystretch}{1}
\label{wmdp dataset}
\setlength{\tabcolsep}{8pt}
\begin{tabular}{c|ccccc|ccccc}
\toprule
\textbf{Dataset} & \multicolumn{5}{c|}{\textbf{WMDP  (\ttsmall{Vicuna-13B})}}                            & \multicolumn{5}{c}{\textbf{MUSE  (\ttsmall{MUSE-7B})}}                                \\ 
\textbf{Metric}  & \textbf{RG↓}   & \textbf{Pr↓}   & \textbf{F1↓}   & \textbf{MU↑}   & \textbf{FRT↑} & \textbf{RG↓}   & \textbf{Pr↓}   & \textbf{F1 ↓}  & \textbf{MU↑}   & \textbf{FRT↑} \\ \midrule
\textbf{GA+GD}   & 82.57          & 35.19          & 76.84          & 55.36          & 0.85          & 82.10          & 20.70          & 76.96          & 38.13          & 0.64          \\
\textbf{DPO+GD}  & 82.42          & 35.57          & 75.37          & 54.14          & 0.84          & 84.73          & 38.94          & 75.50          & 34.19          & 0.51          \\
\textbf{NPO+GD}  & 81.63          & 33.02          & 75.52          & 55.28          & 0.87          & \colorbox{lightblue!50}{81.84}          & 22.94          & 74.63          & 38.00          & 0.64          \\
\textbf{TV}      & 82.95          & 34.70          & 77.36          & 54.64          & 0.84          & 90.60          & \textbf{20.03} & 77.37          & 36.61          & 0.58          \\
\textbf{ULD}     & 81.04          & 32.17          & 76.62          & 55.38          & 0.88          & 89.26          & 31.76          & 74.75          & \textbf{41.51} & 0.64          \\
\textbf{WHP}     & 82.57          & 33.64          & \colorbox{lightblue!50}{74.94}          & 56.25          & 0.89          & 83.20          & 23.90          & \colorbox{lightblue!50}{69.31}          & 38.20          & \colorbox{lightblue!50}{0.65}          \\
\textbf{RMU}     & 83.16          & 38.93          & 78.49          & \textbf{60.14} & 0.90          & 91.92          & 62.16          & 78.89          & 38.80          & 0.50          \\
\textbf{\ours}   & \textbf{80.24} & \textbf{31.83} & \colorbox{lightpurple!50}{\textbf{73.71}} & 56.35          & \textbf{0.91} & \colorbox{lightpurple!50}{\textbf{78.60}} & 21.48          & \colorbox{lightpurple!50}{\textbf{62.39}} & 38.77          & \colorbox{lightpurple!50}{\textbf{0.72}} \\
\bottomrule
\end{tabular}
% \end{scriptsize}
% \vskip -0.2in
\end{table*}

\subsection{Evaluation on Larger Deep Reasoning Models}
\label{Evaluation on Larger Deep Reasoning Models}
It is interesting to observe that models struggle to forget latent one-hop reasoning data (as shown in Tab.~\ref{zsre_rephrase}). Given the growing popularity of deep reasoning models such as OpenAI’s o1 and DeepSeek-R1, it is worth exploring whether these reasoning-enhanced models can achieve better forgetting generalisation on one-hop questions related to the forget data. To investigate this, we conduct experiments on \ttsmall{DeepSeek-R1-Distill-Llama-8B} using the ZSRE One-hop dataset. The model is first fine-tuned using LoRA with rank = 8, after which we apply various unlearning methods, updating only the LoRA parameters. As shown in Tab.~\ref{deep reason}, we have the following key observations:

$\blacktriangleright$  \textbf{Improved unlearning effect with deep reasoning models.} The Forget ROUGE score of \ttsmall{DeepSeek-R1-Distill-Llama-8B} improves by 16.71\% compared to \ttsmall{LLaMA2-7B}. 
However, this improvement comes at the cost of a 25.62\% decrease in model utility.  
This trade-off is expected, as stronger unlearning typically makes it harder to maintain model utility. Moreover, \ours~consistently demonstrates better forgetting performance, with an absolute improvement of up to 2.71\% (57.66 $\rightarrow$ 54.95).

$\blacktriangleright$ \textbf{Forgetting one-hop knowledge remains challenging.} Interestingly, while all methods achieve notable unlearning on the forget set and rephrased forget set, they still struggle with one-hop reasoning data, showing a maximum unlearning gap of 18.87\%.
This result aligns with our observations on \ttsmall{LLaMA2-7B} — one-hop questions serves as challenging latent knowledge to forget.

$\blacktriangleright$ \textbf{Deep reasoning models show better one-hop unlearning performance.} The gap between unlearning vanilla forget data and one-hop data on \ttsmall{LLaMA2-7B} is 23.77\%.  
Compared to \ttsmall{LLaMA2-7B}, the \ttsmall{DeepSeek-R1-Distill-Llama-8B} model exhibits better one-hop forgetting ability, with an improvement of up to 4.9\%. 
This suggests that enhancing a language model's reasoning ability can improve its capacity to unlearn complex, reasoning-based knowledge.

In conclusion, deep reasoning models demonstrate improved one-hop unlearning performance. However, a substantial gap remains between the unlearning of forget data and one-hop data. Moreover, larger models face challenges in preserving utility while achieving effective unlearning, highlighting the need for further investigation in this area.

\begin{table*}[h]
\caption{The impact of the percentage of perturbed tokens $K$ on \ttsmall{Phi-1.3B}. To balance forgetting and retention, we set $K=0.4$ in our main experiments.}
\label{K}
\setlength{\tabcolsep}{8pt}
% \renewcommand{\arraystretch}{1}
% \vskip 0.1in
\begin{center}
% \begin{footnotesize}
\begin{tabular}{c|ccc|ccc|ccc|cc}
\multicolumn{12}{c}{\textbf{\textit{\ttsmall{TOFU Forget01 Dataset}}}} \\
\toprule
\textbf{Split}  & \multicolumn{3}{c|}{\textbf{Forget}}             & \multicolumn{3}{c|}{\textbf{Rephrased Forget}}  & \multicolumn{3}{c|}{\textbf{Retain}}             & \multirow{2}{*}{\textbf{MU↑}} & \multirow{2}{*}{\textbf{FRT↑}} \\
\textbf{Metric} & \textbf{RG↓}   & \textbf{Pr↓}   & \textbf{TR↑}   & \textbf{RG↓}   & \textbf{Pr↓}  & \textbf{TR↑}   & \textbf{RG↑}   & \textbf{Pr↑}   & \textbf{TR↑}   &                               &                                \\ \midrule
K=0.1           & 47.19          & 37.96          & 58.74          & 40.94          & 32.32         & 59.51          & \textbf{82.04} & \textbf{88.47} & \textbf{48.45} & 51.71                         & 1.31                           \\
K=0.2           & 44.98          & 35.66          & 60.50           & 40.20           & 29.35         & 59.83          & 79.18          & 87.27          & 47.96          & 51.55                         & 1.38                           \\
K=0.3           & 43.93          & 32.75          & 61.64          & 39.75          & 28.41         & 61.38          & 80.70          & 87.34          & 47.41          & 51.92                         & 1.43                           \\
\colorbox{gray!40}{K=0.4}           & 43.38          & \colorbox{lightpurple!50}{30.58}          & \colorbox{lightpurple!50}{63.76}          & 39.70           & 27.69         & 61.71          & 79.17          & 87.21          & 47.98          & 51.21                         & \colorbox{lightpurple!50}{1.44}                           \\
K=0.5           & 42.99          & 30.29          & 64.87          & 38.41          & 25.09         & 63.41          & 78.42          & 87.42          & 48.08          & 51.42                         & 1.50                           \\
K=0.6           & 42.85          & 30.02          & 65.96          & 37.63          & 26.71         & 64.05          & 77.77          & 85.76          & 46.67          & 51.55                         & 1.50                           \\
K=0.7           & 42.49          & 30.51          & 66.06          & 36.76          & 26.04         & 65.54          & 77.24          & 85.29          & 46.36          & \colorbox{lightpurple!50}{\textbf{52.17}}                & 1.54                           \\
K=0.8           & 42.19          & 28.39          & 67.52          & 37.00             & 25.21         & 66.56          & 77.73          & 85.36          & 46.31          & 51.82                         & \textbf{1.56}                  \\
K=0.9           & 41.89          & 29.31          & 67.21          & 36.98          & 26.13         & 66.85          & 77.17          & 85.08          & 46.19          & 50.39                         & 1.50                           \\
K=1.0           & \textbf{41.57} & \textbf{27.84} & \textbf{68.46} & \textbf{36.74} & \textbf{24.5} & \textbf{68.08} & 76.28          & 84.63          & 46.24          & 50.24                         & 1.54                           \\ \bottomrule
\end{tabular}
\end{center}
\end{table*}

\subsection{Evaluation on More Diverse Datasets}
\label{Evaluation on More Diverse Datasets}
To further evaluate the unlearning effectiveness, we conduct experiments on more challenging and diverse datasets—WMDP~\cite{WMDP} and MUSE~\cite{MUSE}—where knowledge is often distributed across long-form documents. For WMDP, we select 50 samples from BIO split as the forget set and 350 samples as the retain set. The target model is \ttsmall{Vicuna-13B}, fine-tuned using LoRA with rank = 8. For MUSE, we conduct experiments on BOOK split using \ttsmall{MUSE-books-7B} as the target model. To improve testing efficiency, we randomly sample 40 instances from the verbmem-forget split for unlearning and use the remaining 60 instances for retention. The average input length is 990.3 tokens, with a maximum length of 1031 tokens.

As shown in Tab.~\ref{wmdp dataset}, RMU achieves the highest utility but at the cost of effective unlearning. This may be attributed to RMU's design, which focuses on unlearning with limited samples rather than leveraging the entire forget set—thereby limiting its overall unlearning performance. In contrast, \ours~achieves a stronger forgetting effect, with improvements of 1.23\% on WMDP and 3.24\% on MUSE, and consistently attains the best FRT ratio. These results highlight the effectiveness of \ours~on larger models and longer-context scenarios.

\begin{table}[]
\centering
\footnotesize
\caption{Computational overhead of existing unlearning methods.}
% \vskip 0.1in
\label{overhead}
\begin{tabular}{c|cc}
\toprule
\textbf{Metric} & \textbf{Train step time (seconds)} & \textbf{Memory Usage (MiB)} \\ \midrule
\textbf{GA+KL}     & 22.9342                & 35056                 \\
\textbf{DPO+KL}    & 47.1367                  & 48132                 \\
\textbf{NPO+KL}    & 31.5571                  & 36182                 \\
\textbf{TV}     & 14.2505                  & 20442                 \\
\textbf{ULD}    & 32.4175                  & 31014                 \\
\textbf{WHP}    & 24.1673                  & 20442                 \\
\textbf{\ours}  & 155.7616                  & 52384               \\ 
\textbf{\oursfast}  & 24.3362                  & 39054                \\ \bottomrule
\end{tabular}
\end{table}

\subsection{Computational Overhead}
\label{Computational Overhead}
We evaluated the computational overhead of existing methods from two perspectives: \textbf{training step time} and \textbf{memory usage}. The experiments were conducted on the \ttsmall{Phi-1.3B} model using the TOFU Forget01 dataset, with a batch size of 4 for 5 epochs. We report the total training step time and the maximum memory usage. As shown in Tab.~\ref{overhead}, although \ours~requires more training time, most of the overhead comes from computing the \metric~to identify the top-$K$\% sensitive tokens. Notably, this step can be performed offline prior to training, significantly reducing both time and memory consumption during actual training. Moreover, our proposed fast variant, \oursfast, introduces no additional cost in training step time or memory usage, confirming its computational efficiency.

\subsection{Percentage of Perturbed Tokens $K$}
\label{K ablation}
In \ours, we identify the top-$K$ most sensitive tokens and apply perturbations to them. The choice of $K$ directly influences the unlearning effect. We analyze the impact of varying $K$ from 0.1 to 1.0 in increments of 0.1, where each value represents $K$\% of the tokens to be perturbed, while keeping all other hyper-parameters fixed. 
As shown in Tab.~\ref{K}, increasing $K$ consistently enhances the unlearning ability, with Forget Rouge improved by up to 5.62\% (47.19 $\rightarrow$ 41.57) on the TOFU dataset. % and 11.67\% (63.99 $\rightarrow$ 52.32) on the HP dataset. 
However, this improvement comes at the cost of reduced model utility. This trade-off is expected, as perturbing more tokens makes it harder for the model to recover and correctly predict the unlearning samples. Although the Forget Retain Trade-off (FRT) score peaks when $K=0.8$ to $0.9$, the performance on retained knowledge drops significantly—up to 5.76\% in Rouge (82.04 $\rightarrow$ 76.28) and 3.84\% in Probability (88.47 $\rightarrow$ 84.63). Such degradation may negatively impact the model's utility on preserved knowledge.
To strike a balance between forgetting and retention, we set $K=0.4$ in our main experiments, which achieves effective unlearning while preserving most of the model’s capacity to retain relevant knowledge. %For practical applications, we recommend selecting $K$ in the range of 0.3 to 0.4.

\begin{table*}[]
\caption{The impact of the perturbation ratio $P$. Experiments are conducted on the TOFU Forget01 dataset using \ttsmall{Phi-1.3B}. Different values of $P$ lead to varying trade-offs, we select $P=0.4$ as the optimal perturbation ratio.}
\label{P}
\setlength{\tabcolsep}{12pt}
% \vskip -0.1in
\begin{center}
\begin{small}
\begin{tabular}{c|ccc|ccc|cc}
\toprule
\textbf{Dataset} & \multicolumn{3}{c|}{\textbf{Forget}}       & \multicolumn{3}{c|}{\textbf{Rephrased Forget}}   & \multirow{2}{*}{\textbf{MU↑}} & \multirow{2}{*}{\textbf{FRT↑}} \\
\textbf{Metric}  & \textbf{RG↓} & \textbf{Pr↓} & \textbf{TR↑} & \textbf{RG↓}   & \textbf{Pr↓}   & \textbf{TR↑}   &                               &                                \\ 
\midrule
P=0.1  & 52.12        & 43.00        & 56.18        & 42.28          & 31.00          & 53.79          & 50.90                         & 1.21                           \\
P=0.2  & 50.08        & 37.28        & 57.03        & 40.74          & 27.06          & 57.53          & 50.55                         & 1.30                           \\
P=0.3  & 48.02        & 34.68        & 61.40        & 42.16          & 27.50          & 59.13          & \colorbox{lightpurple!50}{\textbf{51.12}}                & 1.34                           \\
\colorbox{gray!40}{P=0.4}  & 46.26        & 27.11        & 64.65        & 39.95          & 22.76          & 63.44          & 50.14                         & \colorbox{lightpurple!50}{1.47}                           \\
P=0.5  & 47.17        & 30.64        & 62.70        & 41.09          & 24.62          & 62.60          & 48.00                         & 1.34                           \\
P=0.6 & 44.47        & 25.99        & 62.43        & 41.33          & 21.39          & 62.61          & 48.61                         & 1.46                           \\
P=0.7 & 41.69        & 26.95        & 65.19        & 41.10          & 22.84          & 64.38          & 48.79                         & 1.47                           \\
P=0.8  & 42.37        & 24.87        & 65.03        & 39.97          & 21.18          & 64.96          & 47.51                         & 1.48                           \\
P=0.9 & \textbf{41.11}        & 22.49        & 66.79        & 38.76          & 19.08          & 64.88          & 48.08                         & 1.58                           \\
P=1.0  & 41.53        & \colorbox{lightpurple!50}{\textbf{21.10}}      & \colorbox{lightpurple!50}{\textbf{68.66}}        & \textbf{37.11} & \colorbox{lightpurple!50}{\textbf{18.31}} & \textbf{66.21} & 48.61                         & \textbf{1.65}                  \\ 
\bottomrule
\end{tabular}
\end{small}
\end{center}
% \vskip -0.15in
\end{table*}

\begin{table*}[]
\caption{A comparison of different retain loss settings. w/o GDR refers to using the vanilla forget loss, while w/ KLR indicates the combination with KLR. Overall, \oursfast~effectively preserves the model's utility.}
\label{different_retain_loss}
\setlength{\tabcolsep}{12pt}
\begin{center}
\begin{small}
\begin{tabular}{c|ccc|ccc|cc}
\toprule
\textbf{Dataset} & \multicolumn{3}{c|}{\textbf{Forget}}             & \multicolumn{3}{c|}{\textbf{Rephrased Forget}}   & \multirow{2}{*}{\textbf{MU↑}} & \multirow{2}{*}{\textbf{FRT↑}} \\
\textbf{Metric}  & \textbf{RG↓}   & \textbf{Pr↓}   & \textbf{TR↑}   & \textbf{RG↓}   & \textbf{Pr↓}   & \textbf{TR↑}   &                               &                                \\
\midrule
\textbf{\oursfast}    & 44.60          & 47.58          & 67.25          & 32.72          & \textbf{37.99} & 65.96          & \colorbox{lightpurple!50}{\textbf{64.80}}                & \textbf{1.59}                  \\
\textbf{w/o GDR} & 44.88          & 47.52          & 66.44          & 34.23          & 39.67          & 65.50          & \colorbox{lightblue!50}{60.25}                         & 1.45                           \\
\textbf{w/ KLR}  & \textbf{41.54} & \textbf{45.43} & \textbf{67.45} & \textbf{31.92} & 38.15          & \textbf{66.44} & \colorbox{lightblue!50}{60.55}                         & 1.54                               \\ 
\bottomrule
\end{tabular}
\end{small}
\end{center}
\end{table*}

\subsection{Perturbation Ratio $P$}
\label{Perturbation Ratio}
We analyze the impact of the Perturbation Ratio $P$, 
which ranges from $0.1$ to $1.0$ in increments of $0.1$. The experiments are conducted on the TOFU Forget01 dataset using \ttsmall{Phi-1.3B}, keeping other parameters constant. 

As shown in Tab.~\ref{P}, even a small amount of noise ($P=0.1$) is sufficient to achieve a notable unlearning effect. Moreover, increasing the noise ratio further enhances the effectiveness of unlearning. Specifically, when $P=1.0$, almost all metrics on the Forget and Rephrased Forget data achieve their best values. This is expected, as higher noise levels make it harder for the model to recall related facts, resulting in a more fact-unrelated probability distribution and better unlearning performance. 

However, the model utility decreases by up to 3.61\% when $P \geq 0.5$. This suggests that excessive noise can hinder the model's sentence comprehension and increase uncertainty, unintentionally affecting irrelevant knowledge generation. The model utility is the highest when $P=0.3$. Since different values of $P$ result in varying trade-offs, we select $P=0.4$ as the optimal perturbation ratio. This choice is based on the fact that the FRT ratio at $P=0.4$ is higher than at $P=0.3$, while maintaining considerable model utility, indicating a better balance between unlearning performance and model utility. Therefore, we intuitively set $P=0.4$ for all experiments.

\subsection{Different Retain Loss}
\label{Different Retain Loss}
We investigate the impact of different retain loss functions while keeping other parameters fixed. The experiments are conducted on the TOFU Forget10 dataset using \ttsmall{LLaMA2-7B}. \oursfast~typically employs the forget loss combined with GDR, with a retain weight of $RW=1$. Here, w/o GDR refers to using the vanilla forget loss without any retain loss, while w/ KLR denotes the combination of the vanilla forget loss with KLR, applying the same retain weight. As shown in Tab.~\ref{different_retain_loss}, using the vanilla forget loss achieves strong unlearning performance but may slightly impair model utility. While incorporating KLR can improve model utility, the enhancement is less significant compared to using GDR as the retain loss. Therefore, we primarily adopt GDR as the retain loss in our experiments.

\begin{table*}[]
\caption{The impact of the Tuning Coefficient $C$. Experiments are conducted on the TOFU Forget01 dataset using \ttsmall{Phi-1.3B}. Increasing $C$ results in an improved unlearning effect, but at the cost of decreased model utility. To balance model utility with effective unlearning, we select $C=0.1$ for our experiments.}
\label{C}
\setlength{\tabcolsep}{12pt}
% \vskip -0.1in
\begin{center}
\begin{small}
\begin{tabular}{c|ccc|ccc|cc}
\toprule
\textbf{Dataset} & \multicolumn{3}{c|}{\textbf{Forget}}       & \multicolumn{3}{c|}{\textbf{Rephrased Forget}}   & \multirow{2}{*}{\textbf{MU↑}} & \multirow{2}{*}{\textbf{FRT↑}} \\
\textbf{Metric}  & \textbf{RG↓} & \textbf{Pr↓} & \textbf{TR↑} & \textbf{RG↓}   & \textbf{Pr↓}   & \textbf{TR↑}   &                               &      \\ 
\midrule
C=0.0            & 49.69          & 36.45         & 62.02          & 42.08          & 29.45         & 59.81          & \textbf{50.50}                & 1.28                           \\
\colorbox{gray!40}{C=0.1}            & 46.26          & 27.11         & 64.65          & 39.95          & 22.76         & 63.44          & \colorbox{lightpurple!50}{50.14}                         & 1.47                           \\
C=0.2            & 37.62          & 13.66         & 70.43          & 38.76          & 12.18         & 70.10          & 47.73                         & 1.87                           \\
C=0.3            & 24.64          & 2.84          & 71.73          & 25.77          & 2.70          & 72.37          & 39.66                         & 2.84                           \\
C=0.4            & \textbf{12.76} & \textbf{0.83} & 73.13          & \textbf{15.39} & \textbf{0.88} & 72.89          & 31.85                         & \textbf{4.27}                  \\
C=0.5            & 18.11          & 1.38          & 75.68          & 17.16          & 1.36          & 75.04          & 38.81                         & 4.08                           \\
C=0.6            & 19.54          & 1.54          & 77.77          & 21.73          & 1.52          & \textbf{77.60} & 37.56                         & 3.39                           \\
C=0.7            & 18.51          & 1.91          & 78.28          & 20.64          & 1.69          & 77.36          & 37.59                         & 3.52                           \\
C=0.8            & 16.15          & 1.69          & \textbf{78.65} & 19.31          & 1.43          & 77.10          & 37.35                         & 3.87                           \\
C=0.9            & 15.46          & 1.52          & 77.68          & 17.25          & 1.28          & 76.50          & 36.92                         & 4.16                           \\
C=1.0            & 17.64          & 1.48          & 77.19          & 18.66          & 1.24          & 76.21          & 36.66                         & 3.76                  
 \\ 
\bottomrule
\end{tabular}
\end{small}
\end{center}
% \vskip -0.15in
\end{table*}

\subsection{Tuning Coefficient $C$}
\label{Tuning Coefficient}
We investigate the impact of the Tuning Coefficient $C$, which varies from $0.0$ to $1.0$ in increments of $0.1$. The experiments are conducted on the TOFU Forget01 dataset using \ttsmall{Phi-1.3B}, with all other parameters held constant. As shown in Tab.~\ref{C}, the best model utility is achieved when $C=0.0$, where only the corrupted-run probability distribution is used. While this setting maintains high model utility, the unlearning effect is insufficient, as the top-ranked token in the clean-run probability distribution is not fully suppressed. As $C$ increases, the unlearning effect improves, reaching its peak when $C=0.4$. Then it begins to fluctuate as $C$ continues to increase. Correspondingly, model utility decreases with larger values of $C$, which is expected, as higher $C$ values subtract more information from $p(y|x')$, potentially disrupting the distribution of irrelevant knowledge. To balance model utility with effective unlearning, we select $C=0.1$ for our experiments.

\begin{table*}[]
\caption{Experimental results for implementing perturbation at the discrete-token level to subject words, denoted as $\ours_{dis}$. Surprisingly, \oursfastdis~demonstrates a stronger unlearning effect on some datasets, particularly achieving 19.26\% in Forget Rouge and 4.64\% in Probability on the Forget01 dataset using \texttt{LLaMA2-7B}. These results highlight the adaptability of our approach across different dimensions.}
% \vskip -0.15in
\label{Discrete-Token}
\setlength{\tabcolsep}{2.3pt}
\renewcommand{\arraystretch}{1.4}
\begin{center}
\begin{small}
\begin{tabular}{c|c|c|ccc|ccc|ccc|ccc|cc}
\midrule
\multirow{2}{*}{\textbf{Model}} & \multirow{2}{*}{\textbf{TOFU}}     & \textbf{Dataset}                        & \multicolumn{3}{c|}{\textbf{Forget data}}         & \multicolumn{3}{c|}{\textbf{Retain data}}         & \multicolumn{3}{c|}{\textbf{Real Authors}}        & \multicolumn{3}{c|}{\textbf{Real World}}          & \multirow{2}{*}{\textbf{MU↑}} & \multirow{2}{*}{\textbf{FRT↑}} \\
&      & \textbf{Metric}                         & \textbf{RG↓}   & \textbf{Pr↓}   & \textbf{TR↑}   & \textbf{RG↑}   & \textbf{Pr↑}   & \textbf{TR↑}   & \textbf{RG↑}   & \textbf{Pr↑}   & \textbf{TR↑}   & \textbf{RG↑}   & \textbf{Pr↑}   & \textbf{TR↑}   &                               &
\\ \cline{1-17}
\multirow{6}{*}{\textbf{\begin{tabular}[c]{@{}c@{}}Phi\\ (1.3B)\end{tabular}}}  & \multirow{2}{*}{\textbf{Forget01}} 
& \textbf{\oursfast}    & 46.26 & \textbf{27.11} & \textbf{64.65} & 67.74          & 79.50          & 44.96          & 41.90          & 37.85          & 44.04          & \textbf{76.14}          & \textbf{41.75}          & \textbf{50.39}          & 50.14                         & 1.37               \\
 &       & \textbf{\oursfastdis} & \textbf{35.46} & 27.53 & 53.67          & \textbf{84.31} & \textbf{89.36} & \textbf{48.18} & \textbf{46.23} & \textbf{38.26} & \textbf{46.71} & 76.10 & 40.91          & 49.67          & \textbf{52.72}                & \textbf{1.67}                  \\ \cline{2-17}
& \multirow{2}{*}{\textbf{Forget05}} & \textbf{\ours}           & 42.67          & \textbf{25.17} & \textbf{62.96} & \colorbox{lightblue!50}{\textbf{67.97}} & \colorbox{lightblue!50}{\textbf{77.49}} & \textbf{44.79} & 43.23          & \textbf{37.94} & \textbf{45.60} & 76.13          & \textbf{43.10} & \textbf{52.65} & \colorbox{lightpurple!50}{\textbf{50.94}}                & \textbf{1.50}                  \\
&             & \textbf{\oursfastdis} & \textbf{36.94} & 36.76          & 60.86          & \colorbox{lightpurple!50}{42.37}          & \colorbox{lightpurple!50}{58.66}          & 41.05          & \textbf{45.82} & 37.43          & 44.93          & \textbf{78.40} & 40.73          & 48.92          & \colorbox{lightblue!50}{46.50}                         & 1.26                           \\ \cline{2-17}
& \multirow{2}{*}{\textbf{Forget10}} & \textbf{\oursfast}           & 46.55          & \textbf{41.93} & \textbf{57.51} & \colorbox{lightblue!50}{\textbf{81.59}} & \colorbox{lightblue!50}{\textbf{86.17}} & \textbf{47.10} & 35.23          & \textbf{37.64} & \textbf{45.17} & \textbf{75.28} & \textbf{41.25} & \textbf{49.92} & \colorbox{lightpurple!50}{\textbf{50.07}}                & \textbf{1.13}                  \\
&             & \textbf{\oursfastdis} & \textbf{44.90} & 57.06          & 55.92          & \colorbox{lightpurple!50}{46.25}          & \colorbox{lightpurple!50}{63.60}          & 43.77          & \textbf{49.35} & 36.89          & 44.34          & 73.65          & 39.78          & 47.41          & \colorbox{lightblue!50}{47.39}                         & 0.93                           \\ \cline{1-17}
\multirow{6}{*}{\textbf{\begin{tabular}[c]{@{}c@{}}LLaMA\\ (7B)\end{tabular}}} & \multirow{2}{*}{\textbf{Forget01}} & \textbf{\oursfast}           & \colorbox{lightblue!50}{30.42}          & \colorbox{lightblue!50}{16.29}          & \textbf{74.11} & \textbf{86.67} & 88.55          & \textbf{43.51} & \textbf{92.80} & \textbf{51.82} & \textbf{66.34} & \textbf{89.17} & \textbf{49.28} & \textbf{63.02} & \textbf{65.06}                & 2.79                           \\
&     & \textbf{\oursfastdis} & \colorbox{lightpurple!50}{\textbf{19.26}} & \colorbox{lightpurple!50}{\textbf{4.64}}  & 73.92          & 83.72          & \textbf{89.11} & 42.39          & 91.00          & 51.19          & 66.05          & 87.46          & 48.78          & 62.66          & 64.14                         & \textbf{5.37}                  \\
\cline{2-17}
& \multirow{2}{*}{\textbf{Forget05}} & \textbf{\oursfast}           & 33.66          & 39.23          & 64.39          & \textbf{83.63} & \textbf{88.24} & \textbf{41.89} & 91.30          & 52.57          & 68.21          & \textbf{89.60} & 49.77          & 63.94          & \textbf{64.89}                & 1.78                           \\
&        & \textbf{\oursfastdis} & \textbf{30.39} & \textbf{36.55} & \textbf{65.45} & 72.08          & 79.45          & 41.19          & \textbf{91.50} & \textbf{52.61} & \textbf{68.67} & 88.32          & \textbf{50.15} & \textbf{64.52} & 63.38                         & \textbf{1.89}                  \\
\cline{2-17}
& \multirow{2}{*}{\textbf{Forget10}} & \textbf{\oursfast}           & 44.60          & \textbf{47.58} & \textbf{67.25} & \textbf{94.14} & \textbf{94.68} & 41.23          & 90.30          & 51.66          & 66.56          & \textbf{88.75} & 48.48          & 62.10          & \textbf{64.80}                & 1.41                           \\
&            & \textbf{\oursfastdis} & \textbf{40.32} & 48.68          & 62.45          & 70.72          & 79.47          & \textbf{41.56} & \textbf{91.50} & \textbf{53.81} & \textbf{70.03} & 87.89          & \textbf{50.86} & \textbf{65.92} & 63.93         & \textbf{1.44}             \\
\midrule
\end{tabular}
\end{small}
\end{center}
% \vskip -0.15in
\end{table*}

\subsection{Discrete-Token Level Perturbation}
\label{Discrete-Token Level Perturbation}
Apart from adding random noise to the subject token embeddings, \ours~can also be implemented by perturbing the subject words at the discrete-token level, denoted as \oursfastdis.  
In this experiment, we evaluate the unlearning performance of \oursfastdis~on the TOFU dataset while keeping all other parameters constant. The perturbation type is randomly chosen from deleting, altering, or adding letters to the subject words.  

As shown in Tab.~\ref{Discrete-Token}, \oursfastdis~exhibits exceptional unlearning capability, outperforming \oursfast~in Forget ROUGE across all datasets and models. Moreover, \oursfastdis~achieves an absolute 19.26\% Forget ROUGE and 4.64\% Forget Probability on Forget05 when using \ttsmall{LLaMA2-7B}, surpassing \oursfast~by up to 11.16\% (30.42$\rightarrow$19.26) and 11.65\% (16.29$\rightarrow$4.64), respectively, though with a slightly lower model utility of 0.92\%. 
However, when tested on \ttsmall{Phi-1.3B}, \oursfastdis~does not consistently exhibit superior unlearning performance, and the model utility drops by up to 4.44\% (50.94$\rightarrow$46.50) as the number of unlearning samples increases. Despite this, the FRT ratio of \oursfastdis~still outperforms other baselines. 
In summary, perturbing subject words at the discrete-token level can also prevent the model from recalling the fact and generate fact-unrelated probability distributions, thus achieving unlearning. Both embedding-layer and discrete-token-level noise methods can achieve effective unlearning but result in different trade-offs. Given that the knowledge retention ability of \oursfastdis~may decline as the amount of forgotten data increases, we choose to add noise at the embedding layer as a more promising alternative.

\section{Conclusions}
In this paper, we advance LLM-based knowledge forgetting by shifting the focus from superficial forgetting to the more challenging task of implicit knowledge forgetting. To this end, we present a comprehensive evaluation and reveal the poor generalization of existing methods in implicit unlearning. To address this limitation, we propose \ours, which demonstrates strong unlearning performance across multiple real-world scenarios.

In summary, we highlight a promising direction for implicit knowledge forgetting and the need for more thorough and robust unlearning in LLMs.

\section*{Limitations}
While automatic evaluation metrics provide reliable assessments of model performance, we plan to incorporate human evaluations in future work to further strengthen the evaluation process. Although \ours~does not rely on auxiliary models or classifiers, its perturbation process requires computing token-level gradients, which can incur non-trivial overhead in large-scale settings. To mitigate this, we propose a fast implementation variant with reduced training overhead and cost. The exploration of more efficient methods for handling complex unlearning scenarios that involve reasoning is left to future work.

% \appendices

% \ifCLASSOPTIONcompsoc
%   \section*{Acknowledgments}
% \else
% \fi

% \ifCLASSOPTIONcaptionsoff
%   \newpage
% \fi

% \clearpage

\bibliography{unlearning}

\begin{thebibliography}{10}
\providecommand{\url}[1]{#1}
\csname url@samestyle\endcsname
\providecommand{\newblock}{\relax}
\providecommand{\bibinfo}[2]{#2}
\providecommand{\BIBentrySTDinterwordspacing}{\spaceskip=0pt\relax}
\providecommand{\BIBentryALTinterwordstretchfactor}{4}
\providecommand{\BIBentryALTinterwordspacing}{\spaceskip=\fontdimen2\font plus
\BIBentryALTinterwordstretchfactor\fontdimen3\font minus \fontdimen4\font\relax}
\providecommand{\BIBforeignlanguage}[2]{{%
\expandafter\ifx\csname l@#1\endcsname\relax
\typeout{** WARNING: IEEEtran.bst: No hyphenation pattern has been}%
\typeout{** loaded for the language `#1'. Using the pattern for}%
\typeout{** the default language instead.}%
\else
\language=\csname l@#1\endcsname
\fi
#2}}
\providecommand{\BIBdecl}{\relax}
\BIBdecl

\bibitem{LLaMA}
H.~Touvron, T.~Lavril, G.~Izacard, X.~Martinet, M.~Lachaux, T.~Lacroix, B.~Rozi{\`{e}}re, N.~Goyal, E.~Hambro, F.~Azhar, A.~Rodriguez, A.~Joulin, E.~Grave, and G.~Lample, ``Llama: Open and efficient foundation language models,'' \emph{CoRR}, vol. abs/2302.13971, 2023.

\bibitem{GPT-4}
OpenAI, ``{GPT-4} technical report,'' \emph{CoRR}, vol. abs/2303.08774, 2023.

\bibitem{Machine-Unlearning-1}
Y.~Cao and J.~Yang, ``Towards making systems forget with machine unlearning,'' in \emph{2015 {IEEE} Symposium on Security and Privacy, {SP} 2015, San Jose, CA, USA, May 17-21, 2015}.\hskip 1em plus 0.5em minus 0.4em\relax {IEEE} Computer Society, 2015, pp. 463--480.

\bibitem{Data_Deletion}
A.~Ginart, M.~Y. Guan, G.~Valiant, and J.~Zou, ``Making {AI} forget you: Data deletion in machine learning,'' in \emph{Advances in Neural Information Processing Systems 32: Annual Conference on Neural Information Processing Systems 2019, NeurIPS 2019, December 8-14, 2019, Vancouver, BC, Canada}, H.~M. Wallach, H.~Larochelle, A.~Beygelzimer, F.~d'Alch{\'{e}}{-}Buc, E.~B. Fox, and R.~Garnett, Eds., 2019, pp. 3513--3526.

\bibitem{Right_to_be_Forgotten}
D.~Zhang, P.~Finckenberg{-}Broman, T.~Hoang, S.~Pan, Z.~Xing, M.~Staples, and X.~Xu, ``Right to be forgotten in the era of large language models: Implications, challenges, and solutions,'' \emph{CoRR}, vol. abs/2307.03941, 2023.

\bibitem{Knowledge_Unlearning}
N.~Si, H.~Zhang, H.~Chang, W.~Zhang, D.~Qu, and W.~Zhang, ``Knowledge unlearning for llms: Tasks, methods, and challenges,'' \emph{CoRR}, vol. abs/2311.15766, 2023.

\bibitem{DEPN}
X.~Wu, J.~Li, M.~Xu, W.~Dong, S.~Wu, C.~Bian, and D.~Xiong, ``{DEPN:} detecting and editing privacy neurons in pretrained language models,'' in \emph{Proceedings of the 2023 Conference on Empirical Methods in Natural Language Processing, {EMNLP} 2023, Singapore, December 6-10, 2023}, H.~Bouamor, J.~Pino, and K.~Bali, Eds.\hskip 1em plus 0.5em minus 0.4em\relax Association for Computational Linguistics, 2023, pp. 2875--2886.

\bibitem{In-Context-Unlearning}
M.~Pawelczyk, S.~Neel, and H.~Lakkaraju, ``In-context unlearning: Language models as few-shot unlearners,'' in \emph{Forty-first International Conference on Machine Learning, {ICML} 2024, Vienna, Austria, July 21-27, 2024}.\hskip 1em plus 0.5em minus 0.4em\relax OpenReview.net, 2024.

\bibitem{eco}
C.~Y. Liu, Y.~Wang, J.~Flanigan, and Y.~Liu, ``Large language model unlearning via embedding-corrupted prompts,'' in \emph{Advances in Neural Information Processing Systems 38: Annual Conference on Neural Information Processing Systems 2024, NeurIPS 2024, Vancouver, BC, Canada, December 10 - 15, 2024}, A.~Globersons, L.~Mackey, D.~Belgrave, A.~Fan, U.~Paquet, J.~M. Tomczak, and C.~Zhang, Eds., 2024.

\bibitem{grad_ascent}
J.~Jang, D.~Yoon, S.~Yang, S.~Cha, M.~Lee, L.~Logeswaran, and M.~Seo, ``Knowledge unlearning for mitigating privacy risks in language models,'' in \emph{Proceedings of the 61st Annual Meeting of the Association for Computational Linguistics (Volume 1: Long Papers), {ACL} 2023, Toronto, Canada, July 9-14, 2023}, A.~Rogers, J.~L. Boyd{-}Graber, and N.~Okazaki, Eds.\hskip 1em plus 0.5em minus 0.4em\relax Association for Computational Linguistics, 2023, pp. 14\,389--14\,408.

\bibitem{NPO}
R.~Zhang, L.~Lin, Y.~Bai, and S.~Mei, ``Negative preference optimization: From catastrophic collapse to effective unlearning,'' \emph{CoRR}, vol. abs/2404.05868, 2024.

\bibitem{SNAP}
M.~Choi, D.~Rim, D.~Lee, and J.~Choo, ``{SNAP:} unlearning selective knowledge in large language models with negative instructions,'' \emph{CoRR}, vol. abs/2406.12329, 2024.

\bibitem{taskvectors}
G.~Ilharco, M.~T. Ribeiro, M.~Wortsman, L.~Schmidt, H.~Hajishirzi, and A.~Farhadi, ``Editing models with task arithmetic,'' in \emph{The Eleventh International Conference on Learning Representations, {ICLR} 2023, Kigali, Rwanda, May 1-5, 2023}.\hskip 1em plus 0.5em minus 0.4em\relax OpenReview.net, 2023.

\bibitem{ULD}
J.~Ji, Y.~Liu, Y.~Zhang, G.~Liu, R.~R. Kompella, S.~Liu, and S.~Chang, ``Reversing the forget-retain objectives: An efficient {LLM} unlearning framework from logit difference,'' \emph{CoRR}, vol. abs/2406.08607, 2024.

\bibitem{EUL}
J.~Chen and D.~Yang, ``Unlearn what you want to forget: Efficient unlearning for llms,'' in \emph{Proceedings of the 2023 Conference on Empirical Methods in Natural Language Processing, {EMNLP} 2023, Singapore, December 6-10, 2023}, H.~Bouamor, J.~Pino, and K.~Bali, Eds.\hskip 1em plus 0.5em minus 0.4em\relax Association for Computational Linguistics, 2023, pp. 12\,041--12\,052.

\bibitem{PEFT_unlearn}
J.~Zhang, S.~Chen, J.~Liu, and J.~He, ``Composing parameter-efficient modules with arithmetic operations,'' \emph{CoRR}, vol. abs/2306.14870, 2023.

\bibitem{Rethinking}
S.~Liu, Y.~Yao, J.~Jia, S.~Casper, N.~Baracaldo, P.~Hase, X.~Xu, Y.~Yao, H.~Li, K.~R. Varshney, M.~Bansal, S.~Koyejo, and Y.~Liu, ``Rethinking machine unlearning for large language models,'' \emph{CoRR}, vol. abs/2402.08787, 2024.

\bibitem{TOFU}
P.~Maini, Z.~Feng, A.~Schwarzschild, Z.~C. Lipton, and J.~Z. Kolter, ``{TOFU:} {A} task of fictitious unlearning for llms,'' \emph{CoRR}, vol. abs/2401.06121, 2024.

\bibitem{harry_potter}
R.~Eldan and M.~Russinovich, ``Who's harry potter? approximate unlearning in llms,'' \emph{CoRR}, vol. abs/2310.02238, 2023.

\bibitem{SOUL}
J.~Jia, Y.~Zhang, Y.~Zhang, J.~Liu, B.~Runwal, J.~Diffenderfer, B.~Kailkhura, and S.~Liu, ``{SOUL:} unlocking the power of second-order optimization for {LLM} unlearning,'' in \emph{Proceedings of the 2024 Conference on Empirical Methods in Natural Language Processing, {EMNLP} 2024, Miami, FL, USA, November 12-16, 2024}, Y.~Al{-}Onaizan, M.~Bansal, and Y.~Chen, Eds.\hskip 1em plus 0.5em minus 0.4em\relax Association for Computational Linguistics, 2024, pp. 4276--4292.

\bibitem{zsre}
O.~Levy, M.~Seo, E.~Choi, and L.~Zettlemoyer, ``Zero-shot relation extraction via reading comprehension,'' in \emph{Proceedings of the 21st Conference on Computational Natural Language Learning (CoNLL 2017), Vancouver, Canada, August 3-4, 2017}, R.~Levy and L.~Specia, Eds.\hskip 1em plus 0.5em minus 0.4em\relax Association for Computational Linguistics, 2017, pp. 333--342.

\bibitem{Fusion}
\BIBentryALTinterwordspacing
F.~Wan, X.~Huang, D.~Cai, X.~Quan, W.~Bi, and S.~Shi, ``Knowledge fusion of large language models,'' in \emph{The Twelfth International Conference on Learning Representations, {ICLR} 2024, Vienna, Austria, May 7-11, 2024}.\hskip 1em plus 0.5em minus 0.4em\relax OpenReview.net, 2024. [Online]. Available: \url{https://openreview.net/forum?id=jiDsk12qcz}
\BIBentrySTDinterwordspacing

\bibitem{WMDP}
N.~Li, A.~Pan, A.~Gopal, S.~Yue, D.~Berrios, A.~Gatti, J.~D. Li, A.~Dombrowski, S.~Goel, G.~Mukobi, N.~Helm{-}Burger, R.~Lababidi, L.~Justen, A.~B. Liu, M.~Chen, I.~Barrass, O.~Zhang, X.~Zhu, R.~Tamirisa, B.~Bharathi, A.~Herbert{-}Voss, C.~B. Breuer, A.~Zou, M.~Mazeika, Z.~Wang, P.~Oswal, W.~Lin, A.~A. Hunt, J.~Tienken{-}Harder, K.~Y. Shih, K.~Talley, J.~Guan, I.~Steneker, D.~Campbell, B.~Jokubaitis, S.~Basart, S.~Fitz, P.~Kumaraguru, K.~K. Karmakar, U.~K. Tupakula, V.~Varadharajan, Y.~Shoshitaishvili, J.~Ba, K.~M. Esvelt, A.~Wang, and D.~Hendrycks, ``The {WMDP} benchmark: Measuring and reducing malicious use with unlearning,'' in \emph{Forty-first International Conference on Machine Learning, {ICML} 2024, Vienna, Austria, July 21-27, 2024}.\hskip 1em plus 0.5em minus 0.4em\relax OpenReview.net, 2024.

\bibitem{MUSE}
W.~Shi, J.~Lee, Y.~Huang, S.~Malladi, J.~Zhao, A.~Holtzman, D.~Liu, L.~Zettlemoyer, N.~A. Smith, and C.~Zhang, ``{MUSE:} machine unlearning six-way evaluation for language models,'' \emph{CoRR}, vol. abs/2407.06460, 2024.

\bibitem{Patil}
V.~Patil, P.~Hase, and M.~Bansal, ``Can sensitive information be deleted from llms? objectives for defending against extraction attacks,'' in \emph{The Twelfth International Conference on Learning Representations, {ICLR} 2024, Vienna, Austria, May 7-11, 2024}.\hskip 1em plus 0.5em minus 0.4em\relax OpenReview.net, 2024.

\bibitem{Guardrail}
P.~Thaker, Y.~Maurya, and V.~Smith, ``Guardrail baselines for unlearning in llms,'' \emph{CoRR}, vol. abs/2403.03329, 2024.

\bibitem{Chaff}
X.~Hu, D.~Li, B.~Hu, Z.~Zheng, Z.~Liu, and M.~Zhang, ``Separate the wheat from the chaff: Model deficiency unlearning via parameter-efficient module operation,'' in \emph{Thirty-Eighth {AAAI} Conference on Artificial Intelligence, {AAAI} 2024, Thirty-Sixth Conference on Innovative Applications of Artificial Intelligence, {IAAI} 2024, Fourteenth Symposium on Educational Advances in Artificial Intelligence, {EAAI} 2014, February 20-27, 2024, Vancouver, Canada}, M.~J. Wooldridge, J.~G. Dy, and S.~Natarajan, Eds.\hskip 1em plus 0.5em minus 0.4em\relax {AAAI} Press, 2024, pp. 18\,252--18\,260.

\bibitem{RKLD}
B.~Wang, Y.~Zi, Y.~Sun, Y.~Zhao, and B.~Qin, ``{RKLD:} reverse kl-divergence-based knowledge distillation for unlearning personal information in large language models,'' \emph{CoRR}, vol. abs/2406.01983, 2024.

\bibitem{An_embarrassingly_simple}
Z.~Zhang, F.~Wang, X.~Li, Z.~Wu, X.~Tang, H.~Liu, Q.~He, W.~Yin, and S.~Wang, ``Does your {LLM} truly unlearn? an embarrassingly simple approach to recover unlearned knowledge,'' \emph{CoRR}, vol. abs/2410.16454, 2024.

\bibitem{RWKU}
\BIBentryALTinterwordspacing
Z.~Jin, P.~Cao, C.~Wang, Z.~He, H.~Yuan, J.~Li, Y.~Chen, K.~Liu, and J.~Zhao, ``{RWKU:} benchmarking real-world knowledge unlearning for large language models,'' in \emph{Advances in Neural Information Processing Systems 38: Annual Conference on Neural Information Processing Systems 2024, NeurIPS 2024, Vancouver, BC, Canada, December 10 - 15, 2024}, 2024. [Online]. Available: \url{http://papers.nips.cc/paper\_files/paper/2024/hash/b1f78dfc9ca0156498241012aec4efa0-Abstract-Datasets\_and\_Benchmarks\_Track.html}
\BIBentrySTDinterwordspacing

\bibitem{editing_unlearning_iclr}
V.~Patil, P.~Hase, and M.~Bansal, ``Can sensitive information be deleted from llms? objectives for defending against extraction attacks,'' in \emph{The Twelfth International Conference on Learning Representations, {ICLR} 2024, Vienna, Austria, May 7-11, 2024}.\hskip 1em plus 0.5em minus 0.4em\relax OpenReview.net, 2024.

\bibitem{Representation_Misdirection}
\BIBentryALTinterwordspacing
H.~Dang, H.~Thanh{-}Tung, L.~Nguyen, and N.~Inoue, ``Improving the robustness of representation misdirection for large language model unlearning,'' \emph{CoRR}, vol. abs/2501.19202, 2025. [Online]. Available: \url{https://doi.org/10.48550/arXiv.2501.19202}
\BIBentrySTDinterwordspacing

\bibitem{Deep_Unlearning}
\BIBentryALTinterwordspacing
R.~Wu, C.~Yadav, R.~Salakhutdinov, and K.~Chaudhuri, ``Evaluating deep unlearning in large language models,'' \emph{CoRR}, vol. abs/2410.15153, 2024. [Online]. Available: \url{https://doi.org/10.48550/arXiv.2410.15153}
\BIBentrySTDinterwordspacing

\bibitem{Intrinsic}
Y.~Hong, L.~Yu, S.~Ravfogel, H.~Yang, and M.~Geva, ``Intrinsic evaluation of unlearning using parametric knowledge traces,'' \emph{CoRR}, vol. abs/2406.11614, 2024.

\bibitem{acl_bench}
J.~Yao, E.~Chien, M.~Du, X.~Niu, T.~Wang, Z.~Cheng, and X.~Yue, ``Machine unlearning of pre-trained large language models,'' in \emph{Proceedings of the 62nd Annual Meeting of the Association for Computational Linguistics (Volume 1: Long Papers), {ACL} 2024, Bangkok, Thailand, August 11-16, 2024}, L.~Ku, A.~Martins, and V.~Srikumar, Eds.\hskip 1em plus 0.5em minus 0.4em\relax Association for Computational Linguistics, 2024, pp. 8403--8419.

\bibitem{PISTOL}
X.~Qiu, W.~F. Shen, Y.~Chen, N.~Cancedda, P.~Stenetorp, and N.~D. Lane, ``{PISTOL:} dataset compilation pipeline for structural unlearning of llms,'' \emph{CoRR}, vol. abs/2406.16810, 2024.

\bibitem{TULA}
J.~Du, Z.~Wang, and K.~Ren, ``Textual unlearning gives a false sense of unlearning,'' \emph{CoRR}, vol. abs/2406.13348, 2024.

\bibitem{ROME}
K.~Meng, D.~Bau, A.~Andonian, and Y.~Belinkov, ``Locating and editing factual associations in {GPT},'' in \emph{Advances in Neural Information Processing Systems 35: Annual Conference on Neural Information Processing Systems 2022, NeurIPS 2022, New Orleans, LA, USA, November 28 - December 9, 2022}, S.~Koyejo, S.~Mohamed, A.~Agarwal, D.~Belgrave, K.~Cho, and A.~Oh, Eds., 2022.

\bibitem{Editing_Problems}
Y.~Yao, P.~Wang, B.~Tian, S.~Cheng, Z.~Li, S.~Deng, H.~Chen, and N.~Zhang, ``Editing large language models: Problems, methods, and opportunities,'' in \emph{Proceedings of the 2023 Conference on Empirical Methods in Natural Language Processing, {EMNLP} 2023, Singapore, December 6-10, 2023}, H.~Bouamor, J.~Pino, and K.~Bali, Eds.\hskip 1em plus 0.5em minus 0.4em\relax Association for Computational Linguistics, 2023, pp. 10\,222--10\,240.

\bibitem{ROUGE}
C.-Y. Lin, ``Rouge: A package for automatic evaluation of summaries,'' in \emph{Text summarization branches out}, 2004, pp. 74--81.

\bibitem{DPO}
R.~Rafailov, A.~Sharma, E.~Mitchell, C.~D. Manning, S.~Ermon, and C.~Finn, ``Direct preference optimization: Your language model is secretly a reward model,'' in \emph{Advances in Neural Information Processing Systems 36: Annual Conference on Neural Information Processing Systems 2023, NeurIPS 2023, New Orleans, LA, USA, December 10 - 16, 2023}, A.~Oh, T.~Naumann, A.~Globerson, K.~Saenko, M.~Hardt, and S.~Levine, Eds., 2023.

\bibitem{NPO_Rethinking}
C.~Fan, J.~Liu, L.~Lin, J.~Jia, R.~Zhang, S.~Mei, and S.~Liu, ``Simplicity prevails: Rethinking negative preference optimization for {LLM} unlearning,'' \emph{CoRR}, vol. abs/2410.07163, 2024.

\bibitem{QUARK}
X.~Lu, S.~Welleck, J.~Hessel, L.~Jiang, L.~Qin, P.~West, P.~Ammanabrolu, and Y.~Choi, ``{QUARK:} controllable text generation with reinforced unlearning,'' in \emph{Advances in Neural Information Processing Systems 35: Annual Conference on Neural Information Processing Systems 2022, NeurIPS 2022, New Orleans, LA, USA, November 28 - December 9, 2022}, S.~Koyejo, S.~Mohamed, A.~Agarwal, D.~Belgrave, K.~Cho, and A.~Oh, Eds., 2022.

\bibitem{Models_Sensitivity}
M.~Sclar, Y.~Choi, Y.~Tsvetkov, and A.~Suhr, ``Quantifying language models' sensitivity to spurious features in prompt design or: How {I} learned to start worrying about prompt formatting,'' in \emph{The Twelfth International Conference on Learning Representations, {ICLR} 2024, Vienna, Austria, May 7-11, 2024}.\hskip 1em plus 0.5em minus 0.4em\relax OpenReview.net, 2024.

\bibitem{SSS}
H.~Wang, H.~Sun, J.~Wang, Q.~Qi, Z.~Xia, M.~Zhang, and J.~Liao, ``{SSS:} editing factual knowledge in language models towards semantic sparse space,'' in \emph{Findings of the Association for Computational Linguistics, {ACL} 2024, Bangkok, Thailand and virtual meeting, August 11-16, 2024}, L.~Ku, A.~Martins, and V.~Srikumar, Eds.\hskip 1em plus 0.5em minus 0.4em\relax Association for Computational Linguistics, 2024, pp. 5559--5570.

\bibitem{FIM}
C.~Zhao, P.~T. Fletcher, M.~Yu, Y.~Peng, G.~Zhang, and C.~Shen, ``The adversarial attack and detection under the fisher information metric,'' in \emph{{AAAI}2019, {IAAI} 2019, {EAAI} 2019}.\hskip 1em plus 0.5em minus 0.4em\relax {AAAI} Press, 2019, pp. 5869--5876.

\bibitem{f-Divergence}
Y.~Wen, Z.~Li, W.~Du, and L.~Mou, ``f-divergence minimization for sequence-level knowledge distillation,'' in \emph{Proceedings of the 61st Annual Meeting of the Association for Computational Linguistics (Volume 1: Long Papers), {ACL} 2023, Toronto, Canada, July 9-14, 2023}, A.~Rogers, J.~L. Boyd{-}Graber, and N.~Okazaki, Eds.\hskip 1em plus 0.5em minus 0.4em\relax Association for Computational Linguistics, 2023, pp. 10\,817--10\,834.

\bibitem{Hurt}
J.~Gu, H.~Xu, J.~Ma, P.~Lu, Z.~Ling, K.~Chang, and N.~Peng, ``Model editing can hurt general abilities of large language models,'' \emph{CoRR}, vol. abs/2401.04700, 2024.

\bibitem{Recall_of_Factual}
M.~Geva, J.~Bastings, K.~Filippova, and A.~Globerson, ``Dissecting recall of factual associations in auto-regressive language models,'' in \emph{Proceedings of the 2023 Conference on Empirical Methods in Natural Language Processing, {EMNLP} 2023, Singapore, December 6-10, 2023}, H.~Bouamor, J.~Pino, and K.~Bali, Eds.\hskip 1em plus 0.5em minus 0.4em\relax Association for Computational Linguistics, 2023, pp. 12\,216--12\,235.

\bibitem{Dissecting}
\BIBentryALTinterwordspacing
------, ``Dissecting recall of factual associations in auto-regressive language models,'' in \emph{Proceedings of the 2023 Conference on Empirical Methods in Natural Language Processing, {EMNLP} 2023, Singapore, December 6-10, 2023}, H.~Bouamor, J.~Pino, and K.~Bali, Eds.\hskip 1em plus 0.5em minus 0.4em\relax Association for Computational Linguistics, 2023, pp. 12\,216--12\,235. [Online]. Available: \url{https://doi.org/10.18653/v1/2023.emnlp-main.751}
\BIBentrySTDinterwordspacing

\bibitem{entropies}
Y.~Zhang, M.~Galley, J.~Gao, Z.~Gan, X.~Li, C.~Brockett, and B.~Dolan, ``Generating informative and diverse conversational responses via adversarial information maximization,'' in \emph{Advances in Neural Information Processing Systems 31: Annual Conference on Neural Information Processing Systems 2018, NeurIPS 2018, December 3-8, 2018, Montr{\'{e}}al, Canada}, S.~Bengio, H.~M. Wallach, H.~Larochelle, K.~Grauman, N.~Cesa{-}Bianchi, and R.~Garnett, Eds., 2018, pp. 1815--1825.

\end{thebibliography}
\bibliographystyle{IEEEtran}

% \end{IEEEbiography}



\end{document}